\documentclass[letterpaper]{article} 
\usepackage{aaai2027}  
\usepackage[hyphens]{url}  
\usepackage{graphicx} 
\usepackage{natbib}  
\usepackage{caption} 
\usepackage{algorithm}
\usepackage{algorithmic}

\usepackage{newfloat}
\usepackage{listings}
\usepackage{amssymb} 
\usepackage{amsmath} 
\usepackage{cleveref}
\usepackage{multirow}
\usepackage[table]{xcolor}
\usepackage[most]{tcolorbox}
\DeclareCaptionStyle{ruled}{labelfont=normalfont,labelsep=colon,strut=off} 
\floatstyle{ruled}
\newfloat{listing}{tb}{lst}{}
\floatname{listing}{Listing}

\usepackage{booktabs}

\nocopyright

\title{Mitigating Identity Essentialism in LLM Agents with Longitudinal Life Trajectories}
\author{
    Hexi Wang\textsuperscript{\rm 1,2},
    Yujia Zhou\corresponding\textsuperscript{\rm 2,1},
    Bangde Du\textsuperscript{\rm 1},
    Weihang Su\textsuperscript{\rm 1},
    Xinyuan Cao\textsuperscript{\rm 1},
    Qingyi Pan\textsuperscript{\rm 1},\\
    Qingyao Ai\corresponding\textsuperscript{\rm 2,1},
    Yueyue Wu\textsuperscript{\rm 1},
    Min Zhang\textsuperscript{\rm 1},
    Yiqun Liu\textsuperscript{\rm 1}
}
\affiliations{
    \textsuperscript{\rm 1}Department of Computer Science and Technology, Tsinghua University\\
    \textsuperscript{\rm 2}Quan Cheng Laboratory

    whx25@mails.tsinghua.edu.cn, zhouyujia@mail.tsinghua.edu.cn, aiqy@tsinghua.edu.cn
}

\begin{document}

\maketitle

\begin{abstract}
Large language models (LLMs) offer a scalable approach to social simulation, but their credibility depends on how agents are constructed. 
Existing methods can partially reproduce population-level patterns, yet often fail to capture human-like diversity. Our analysis shows that static-profile agents exhibit stronger demographic separation and within-group compression than humans, a pattern consistent with identity essentialism: \textit{demographic labels can encourage models to treat group-average tendencies as individual traits}, homogenizing responses within groups. 
We argue that this limitation arises from two related factors: sparse, static agent representations and the limited ability of prompt-only memory to persistently integrate experience.
Inspired by complementary memory systems, we propose LifeMem, a longitudinal memory framework that combines structured life-event retrieval with agent-specific parametric memory for experience integration. 
Experiments on Understanding Society with three LLMs show that LifeMem improves alignment with human data in terms of response distributions, overall and within-group diversity, and patterns of within-person response change across life stages.
These findings highlight the value of longitudinal life-event memory for constructing more faithful and dynamically evolving social agents.
\end{abstract}

\begin{links}
    \link{Code}{https://github.com/halsayxi/LifeMem}
\end{links}

\section{Introduction}
\label{sec:introduction}

Large language models (LLMs) provide a scalable foundation for social simulation, enabling researchers to study human opinions, attitudes, behaviors, and interactions when conventional surveys or experiments are costly or difficult \citep{aher2023simulate,argyle2023out,park2023generative}. The credibility of such simulations depends critically on how agents are constructed.

\begin{figure*}[t]
    \centering
    \includegraphics[width=\textwidth]{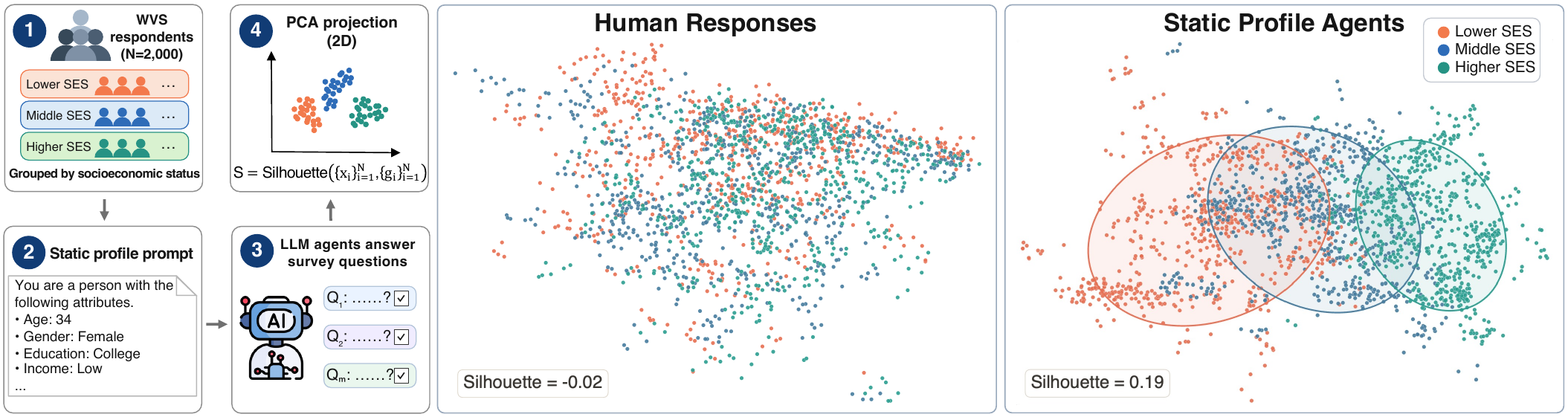}
    \caption{Static demographic conditioning produces a pattern consistent with identity essentialism. We divide 2,000 randomly sampled WVS wave 7 respondents \citep{haerpfer2022world} into three socioeconomic status groups and compare human responses with those of profile-conditioned Llama-8B agents. Each individual's responses are concatenated and projected into a two-dimensional PCA space. The separate projections show greater within-group variation and overlap among humans (silhouette $-0.02$), whereas agents exhibit stronger within-group compression and between-group separation (silhouette $0.19$). Silhouette scores are computed in the original response space before PCA projection.}
    \label{fig:identity-essentialism}
\end{figure*}

Most approaches instantiate agents from demographic labels, personas, or static profiles \citep{argyle2023out,santurkar2023whose,bisbee2024synthetic,hu2024quantifying}. Although these representations effectively condition models on individual attributes, they often produce responses that are less diverse than human data \citep{bisbee2024synthetic,wang2024flatten,xie2026statistical,wang2026psii}. This diversity collapse may reflect a form of \textbf{identity essentialism} \citep{wang2024flatten}. 
In social psychology, essentialism treats social categories as reflecting stable underlying essences, making members of the same category appear fundamentally alike \citep{prentice2007essentialism,bastian2006essentialism}. In social simulation, a corresponding pattern arises when static identity labels become overly predictive of agent responses, compressing within-group variation while amplifying between-group differences. Figure~\ref{fig:identity-essentialism} presents a response-space analysis consistent with this pattern. 
Such distortions may exaggerate demographic differences, reinforce stereotypes, and bias conclusions drawn from simulated populations.

We attribute this limitation to both the content and form of agent memory. First, demographic attributes provide only a sparse and static account of an individual, omitting the richer, evolving experiences associated with between-person differences and within-person change \citep{elder1998lifecourse,curran2011disaggregation}. Second, prompt-based conditioning may lead to diversity collapse \citep{wang2026psii}. As shown in Figure~\ref{fig:event-rag-topk}, explicit memory alone does not ensure that accumulated experience is persistently integrated into the agent.

Inspired by cognitive science, we propose \textbf{LifeMem}, a longitudinal memory framework for constructing dynamically evolving LLM social agents. Human memory is commonly understood as relying on complementary systems: the hippocampal system rapidly preserves individual experiences, whereas the neocortex gradually integrates information across experiences into persistent representations \citep{mcclelland1995complementary,kumaran2016learning}. LifeMem translates this distinction into two complementary components for representing life experiences: (1) structured life-event memory, which explicitly preserves event content, timing, and supporting evidence for retrieval; and (2) agent-specific LoRA adapters, which encode accumulated experiences as parametric memory persistent across questions and updated over time. By combining richer longitudinal information with persistent parametric memory, LifeMem targets limitations in both the content and representation of agent memory, aiming to construct agents that more faithfully reflect individual differences and temporal change.

We evaluate LifeMem on Understanding Society \citep{buck2012understanding} using three instruction-tuned LLMs from different families:
\textbf{Llama-3.1-8B-Instruct} (hereafter, \textbf{Llama-8B}) \citep{grattafiori2024llama},
\textbf{Ministral-3-8B-Instruct-2512} \citep{liu2026ministral3},
and \textbf{Qwen3.5-9B} \citep{qwen2026qwen35}.
Compared to static conditioning, diversity-oriented prompting, non-parametric memory, and control baselines, LifeMem improves alignment with human data in terms of response distributions, overall and within-group diversity, and within-person response change across life stages.
Our contributions are:

\begin{itemize}
\item We identify identity essentialism in static-profile LLM agents, which reduces within-group diversity and exaggerates between-group differences.
\item We propose LifeMem, combining structured life-event memory with agent-specific parametric memory to model diverse and evolving individuals.
\item Across Understanding Society and three LLMs, LifeMem improves alignment with human distributions, diversity, and within-person changes.
\end{itemize}

\section{Related Work}
\label{sec:related_work}

\textbf{Current LLM-based social simulation often relies on persona conditioning and static agent representations.} LLMs are increasingly used as experimental participants and survey respondents \citep{aher2023simulate,argyle2023out,santurkar2023whose,bisbee2024synthetic,hu2024quantifying,lutz2025prompt}. Existing approaches often construct agents from interviews, interaction histories, or social-media records \citep{park2024thousand,du2025simvbg,li2026spirit,du2026twinvoice,guo2026individualturing}, supporting partial alignment with populations. 
However, persona conditioning may reduce response variance or introduce bias \citep{bisbee2024synthetic,boelaert2025machine,wang2024flatten}. 
Moreover, static profiles can compress individuals into demographic types \citep{xie2026statistical,hu2024quantifying}. This pattern resembles identity essentialism, which attributes stable properties to social categories \citep{prentice2007essentialism,bastian2006essentialism}, although demographic conditioning is not inherently essentialist \citep{wang2024flatten}. 
More broadly, LLM social agents are typically fixed rather than updated from temporally ordered observations. In contrast, longitudinal research distinguishes between-person differences from within-person change and uses detailed life-event histories to better characterize individual trajectories \citep{curran2011disaggregation,elder1998lifecourse}.

\begin{figure*}[t]
    \centering
    \includegraphics[width=\textwidth]{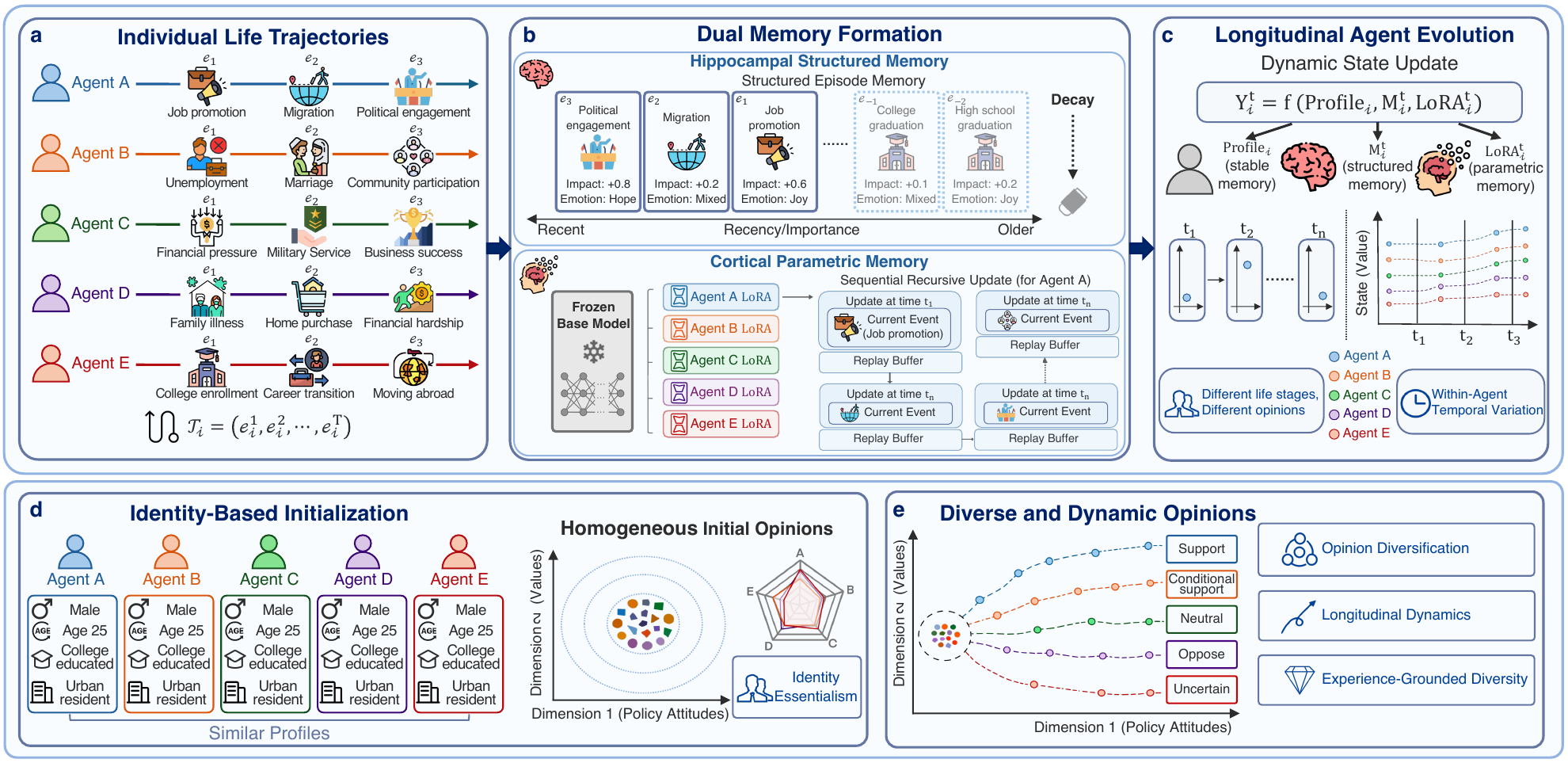}
    \caption{Overview of LifeMem. (a) Individual life trajectories contain diverse longitudinal experiences. (b) Structured memory supports explicit retrieval, while agent-specific LoRA adapters consolidate experiences into parametric memory through recursive updates. (c) The resulting agent state evolves across life stages. (d) Agents are initialized from similar demographic profiles, which may produce homogeneous opinions. (e) LifeMem produces diverse and dynamic opinions.}
    \label{fig:framework}
\end{figure*}

\textbf{Prior work improves diversity at the output, prompt, memory, or parameter level.}
Output-level methods modify sampling, decoding \citep{holtzman2020curious,platt1999probabilistic,li2016diversity,vijayakumar2018diverse,chung2023increasing,wong2026simplestrat,zhang2025cultivating}, or human involvement \citep{abels2025wisdom} to increase response diversity, but the resulting variation is often stochastic rather than grounded in individual experience.
Prompt-level methods introduce multilingual, anti-stereotype, biography-based, population-aligned, or ensemble personas \citep{wang2025multilingual,sivakumar2025bias,lutz2025prompt,ge2024personahub,hu2025populationaligned,ashkinaze2025plurals}, but they generally lack an evolving representation of individuals.
Memory methods store, retrieve, update, or forget experience \citep{maharana2024evaluating,lewis2020rag,park2023generative,zhong2024memorybank,long2026seeing,guo2026memeyevisualcentricevaluationframework}, but retrieval alone may not support persistent integration.
Parameter-level methods encode knowledge or identity into model parameters or activations \citep{su2025parametricrag,chen2025personavectors,wang2026psii}, but are not typically updated from longitudinal observations.

\section{Methodology}
\label{sec:methodology}

This section formulates the longitudinal simulation task and introduces LifeMem's dual-memory framework. We then describe its structured memory for explicit event retrieval and parametric memory for integrating experiences across waves.

\subsection{Problem Formulation}
\label{sec:problem-formulation}

Let $\mathcal{I}=\{1,\ldots,N\}$ denote a population of individuals and $\mathcal{W}=\{1,\ldots,T\}$ the ordered longitudinal waves. 
Each individual $i$ has a static profile $P_i$ constructed from background information and a sequence of life events observed at wave $t$,
\begin{equation} \mathcal{E}_{i,t} = \{e_{i,t,k}\}_{k=1}^{K_{i,t}}, \end{equation}
where $e_{i,t,k}$ denotes the $k$-th observed life event and $K_{i,t}=|\mathcal{E}_{i,t}|$.
The accumulated trajectory through wave $t$ is
\begin{equation}
\mathcal{T}_{i,\leq t}
=
\bigcup_{\tau=1}^{t}\mathcal{E}_{i,\tau}.
\end{equation}

For an evaluation question $q_{t,j}$, LifeMem predicts
\begin{equation}
\hat{y}_{i,t,j}
=
F_{\theta,A_{i,t}}
\left(
P_i,
R_{i,t}(q_{t,j}),
q_{t,j}
\right),
\label{eq:lifemem-prediction}
\end{equation}
where $\theta$ is the frozen base LLM, $A_{i,t}$ is an agent-specific LoRA adapter, and $R_{i,t}(q_{t,j})$ contains events retrieved from structured memory. LifeMem combines question-dependent evidence with a persistent individual state to model between-person differences and within-person change.

\subsection{LifeMem Overview}
\label{sec:lifemem-overview}

LifeMem is a dual-memory framework inspired by complementary learning systems theory, which distinguishes rapid encoding of specific experiences in the hippocampal system from gradual integration in the neocortex \citep{mcclelland1995complementary,oreilly2002hippocampal,oreilly2014complementary}. 
In LifeMem, structured memory stores explicit and traceable life events, while agent-specific LoRA adapters integrate accumulated experiences into persistent parametric memory, as illustrated in Figure~\ref{fig:framework}.

Both memories receive the same longitudinal event stream but serve complementary roles. Structured memory provides relevant evidence for the question, whereas parametric memory integrates accumulated experience across questions and waves without repeatedly adding the full trajectory into the prompt. Temporal decay prioritizes retrieval, and replay preserves earlier information during sequential adapter updates.

\subsection{Hippocampal Structured Memory}
\label{sec:structured-memory}

Each life event $e_{i,t,k}$ is derived from a survey question--answer pair and converted into a second-person statement $x_{i,t,k}$ while retaining its original question and response. A frozen encoder $g_{\phi}$ maps the statement to an embedding
\begin{equation}
\mathbf{h}_{i,t,k}=g_{\phi}(x_{i,t,k}).
\end{equation}
The structured memory stores each event together with its embedding and wave index:
\begin{equation}
\mathcal{M}^{S}_{i,t}
=
\left\{
\left(e_{i,\tau,k},\mathbf{h}_{i,\tau,k},\tau\right)
\mid
1\leq\tau\leq t
\right\}.
\end{equation}
New events are appended without removing earlier records, preserving a traceable account of the observed trajectory.

For a question $q$, retrieval combines semantic relevance with temporal decay:
\begin{equation}
\operatorname{Score}_{t}(q,e_{i,\tau,k})
=
\operatorname{sim}
\left(
g_{\phi}(q),\mathbf{h}_{i,\tau,k}
\right)
\exp[-\lambda(t-\tau)],
\label{eq:temporal-forgetting}
\end{equation}
where $\lambda\geq0$ controls the preference for recent events. This decay modifies retrieval priority rather than deleting older experiences, which remain available when sufficiently relevant. The top-$K$ events form the retrieved evidence:
\begin{equation}
R_{i,t}(q)
=
\operatorname{Top-K}_{e\in\mathcal{M}^{S}_{i,t}}
\operatorname{Score}_{t}(q,e).
\label{eq:event-retrieval}
\end{equation}
The selected statements are inserted into the inference prompt, providing question-dependent evidence without exposing the complete history.

\subsection{Cortical Parametric Memory}
\label{sec:parametric-memory}

LifeMem maintains an independently updated LoRA adapter for each agent while keeping the base LLM frozen. For a target layer $l$, the agent-specific weight is
\begin{equation}
\mathbf{W}_{i,t}^{(l)}
=
\mathbf{W}_{0}^{(l)}
+
\frac{\alpha_{\mathrm{LoRA}}}{r}
\mathbf{B}_{i,t}^{(l)}
\mathbf{A}_{i,t}^{(l)},
\label{eq:lora-update}
\end{equation}
where $r$ and $\alpha_{\mathrm{LoRA}}$ denote the LoRA rank and scaling coefficient. We use $\Theta_{i,t}^{\mathrm{LoRA}}$ to denote all agent-specific LoRA parameters, including $\{\mathbf{A}_{i,t}^{(l)},\mathbf{B}_{i,t}^{(l)}\}_{l}$. Separate adapters allow individuals with similar demographic profiles to develop different parametric states as their trajectories diverge.

Each life event yields two training instances: a survey question--answer pair preserving the observed response and an event-reconstruction instance linking the event to the agent's first-person representation. Together, they encode the survey signal and individualized semantics.

Adapters are updated sequentially across waves, with each agent maintaining an independent LoRA adapter. At the first wave, all adapters are copied from the same PEFT default initialization, where A is randomly initialized and B is zero-initialized, yielding a zero initial LoRA update; thereafter, each adapter is trained independently:
\begin{equation}
\Theta_{i,t}^{\mathrm{LoRA},(0)}
=
\Theta_{i,t-1}^{\mathrm{LoRA}},
\qquad t>1,
\end{equation}
At each wave, training combines instances from current events with a small sample of historical instances sampled uniformly at random for replay:
\begin{equation}
\mathcal{L}_{i,t}
=
\sum_{z\in\mathcal{B}_{i,t}^{\mathrm{cur}}}
\ell(z;\theta,\Theta_{i,t}^{\mathrm{LoRA}})
+
\eta
\sum_{z\in\mathcal{B}_{i,t}^{\mathrm{rep}}}
\ell(z;\theta,\Theta_{i,t}^{\mathrm{LoRA}}),
\label{eq:adapter-objective}
\end{equation}
where $\eta$ controls the contribution of replayed experiences. The base model remains frozen, and only the agent-specific low-rank parameters $\Theta_{i,t}^{\mathrm{LoRA}}$ are optimized. Replay serves as a lightweight mechanism for mitigating catastrophic forgetting rather than a separate memory component.

\section{Experimental Setup}
\label{sec:experimental-setup}

\subsection{Datasets}
\label{sec:datasets}

We evaluate LifeMem on fifteen waves of the UK Household Longitudinal Study (Understanding Society). To ensure consistent longitudinal tracking, we retain only respondents observed in all selected waves.

Survey variables are semantically categorized into three types: demographic attributes for initial profiles, life events describing evolving experiences, and evaluation targets. To prevent leakage, evaluation targets are excluded from both demographic and life-event variables. To evaluate within-person opinion changes, we further identify a shared set of evaluation questions available in every wave of Understanding Society and use them as the longitudinal evaluation set.

\subsection{Baselines}
\label{sec:baselines}

Static conditioning includes \textsc{Direct} and \textsc{Profile}
\citep{argyle2023out,santurkar2023whose,bisbee2024synthetic,hu2024quantifying};
diversity-oriented prompting includes \textsc{Multilingual}
\citep{wang2025multilingual} and \textsc{Anti-Stereotype}
\citep{sivakumar2025bias}; and non-parametric memory includes
\textsc{SimVBG} \citep{du2025simvbg}, \textsc{Full History}
\citep{maharana2024evaluating}, and \textsc{Event RAG}
\citep{lewis2020rag,maharana2024evaluating}. \textsc{Random Event} uses random events to test the importance of respondent-specific trajectories.

\subsection{Evaluation Metrics}
\label{sec:evaluation-metrics}

We use four lower-is-better metrics. 
\emph{KL divergence} measures alignment between human and agent response distributions. 
The \emph{within-group pairwise distance gap} is the absolute difference between human and agent within-group pairwise response distances. A smaller gap indicates that simulated within-group diversity is closer to that of humans, suggesting less identity-essentialist behavior.
The \emph{normalized entropy gap} is the absolute difference between the normalized response entropies of humans and agents.
The \emph{transition-distribution JS divergence} measures how closely the model reproduces the human population-level distribution of response transitions between adjacent survey waves.

\subsection{Implementation Details}
\label{sec:implementation-details}

We use three similarly sized LLMs: Llama-3.1-8B-Instruct, Ministral-3-8B-Instruct-2512, and Qwen3.5-9B. Using a random seed of 42, we sample 100 respondents. All methods share the same samples, questions, backbones, and deterministic decoding settings, with temperature $0$ and sampling disabled. \textsc{Event RAG} uses the stronger bge-m3 retriever, whereas LifeMem uses the lightweight all-MiniLM-L6-v2 encoder. Both methods retrieve the top-$K=5$ events. LifeMem uses a temporal decay factor of $0.105$, rank-$8$ LoRA adapters, replay size $4$, and replay weight $0.5$. Experiments run on one NVIDIA A800-SXM4 80\,GB GPU.

\section{Experiments and Results}
\label{sec:experiments-results}

\begin{table*}[t]
\centering
\setlength{\tabcolsep}{4pt}
\begin{tabular}{l|l|cccc}
\toprule
\textbf{Category}
&
\textbf{Method}
&
KL Divergence $\downarrow$&
Within-Group Gap $\downarrow$&
Entropy Gap $\downarrow$&
Transition JS $\downarrow$
\\
\midrule

\multicolumn{6}{c}{\textbf{\textsc{Llama-3.1-8B-Instruct}}}\\
\midrule

\multirow{2}{*}{\textit{Static Conditioning}}
& Direct
& 15.3734$^{*}$ & 0.5561$^{*}$ & 0.7076$^{*}$ & 0.4580$^{*}$
\\
& Profile
& 7.1522$^{*}$ & 0.3835$^{*}$ & 0.4972$^{*}$ & 0.3886$^{*}$
\\
\multirow{2}{*}{\textit{Diversity-Oriented Prompting}}
& Multilingual
& 11.7367$^{*}$ & 0.3446$^{*}$ & 0.4763$^{*}$ & 0.4570$^{*}$
\\
& Anti-Stereotype
& 8.1485$^{*}$ & 0.4066$^{*}$ & 0.5120$^{*}$ & 0.3592$^{*}$
\\
\multirow{3}{*}{\textit{Non-Parametric Memory}}
& SimVBG
& 6.2912$^{*}$ & 0.3544$^{*}$ & 0.4730$^{*}$ & 0.3709$^{*}$
\\
& Full History
& 4.9344$^{*}$ & 0.3198$^{*}$ & 0.4004$^{*}$ & 0.3633$^{*}$
\\
& Event RAG
& 5.2935$^{*}$ & 0.3163$^{*}$ & 0.3991$^{*}$ & 0.3648$^{*}$
\\
\textit{Control Baseline}
& Random Event
& 5.5673$^{*}$ & 0.3429$^{*}$ & 0.4539$^{*}$ & 0.3803$^{*}$
\\
\rowcolor{gray!20}
\textit{Proposed Method}
& LifeMem
& \textbf{3.4529} & \textbf{0.2886} & \textbf{0.3742} & \textbf{0.3331}
\\
\midrule
\multicolumn{6}{c}{\textbf{\textsc{Ministral-3-8B-Instruct-2512}}}\\
\midrule

\multirow{2}{*}{\textit{Static Conditioning}}
& Direct
& 15.4925$^{*}$ & 0.5648$^{*}$ & 0.7182$^{*}$ & 0.5076$^{*}$
\\
& Profile
& 6.8298$^{*}$ & 0.3622$^{*}$ & 0.4622$^{*}$ & 0.3609
\\
\multirow{2}{*}{\textit{Diversity-Oriented Prompting}}
& Multilingual
& 13.1824$^{*}$ & 0.3942$^{*}$ & 0.5350$^{*}$ & 0.4889$^{*}$
\\
& Anti-Stereotype
& 8.7660$^{*}$ & 0.3946$^{*}$ & 0.5066$^{*}$ & 0.3757
\\
\multirow{3}{*}{\textit{Non-Parametric Memory}}
& SimVBG
& 4.8323$^{*}$ & 0.3272$^{*}$ & 0.4156$^{*}$ & 0.3610
\\
& Full History
& 3.8053$^{*}$ & 0.2927$^{*}$ & 0.3413$^{*}$ & 0.3528
\\
& Event RAG
& 4.3342$^{*}$ & 0.2887$^{*}$ & 0.3413$^{*}$ & 0.4693$^{*}$
\\
\textit{Control Baseline}
& Random Event
& 4.7275$^{*}$ & 0.3156$^{*}$ & 0.4048$^{*}$ & 0.3924
\\
\rowcolor{gray!20}
\textit{Proposed Method}
& LifeMem
& \textbf{1.9659} & \textbf{0.2277} & \textbf{0.2659} & \textbf{0.3399}
\\
\midrule
\multicolumn{6}{c}{\textbf{\textsc{Qwen3.5-9B}}}\\
\midrule

\multirow{2}{*}{\textit{Static Conditioning}}
& Direct
& 15.9998$^{*}$ & 0.5559$^{*}$ & 0.7080$^{*}$ & 0.4452$^{*}$
\\
& Profile
& 5.3214$^{*}$ & 0.3463$^{*}$ & 0.4325$^{*}$ & 0.3471$^{*}$
\\
\multirow{2}{*}{\textit{Diversity-Oriented Prompting}}
& Multilingual
& 10.9903$^{*}$ & 0.3438$^{*}$ & 0.4728$^{*}$ & 0.4405$^{*}$
\\
& Anti-Stereotype
& 6.0185$^{*}$ & 0.3422$^{*}$ & 0.4254$^{*}$ & 0.3356$^{*}$
\\
\multirow{3}{*}{\textit{Non-Parametric Memory}}
& SimVBG
& 4.4864$^{*}$ & 0.3182$^{*}$ & 0.3889$^{*}$ & 0.3298$^{*}$
\\
& Full History
& 3.7659$^{*}$ & 0.2979$^{*}$ & 0.3441$^{*}$ & 0.3203$^{*}$
\\
& Event RAG
& 3.5507$^{*}$ & 0.2673$^{*}$ & 0.3070 & 0.3394$^{*}$
\\
\textit{Control Baseline}
& Random Event
& 3.7848$^{*}$ & 0.3031$^{*}$ & 0.3707$^{*}$ & 0.3466$^{*}$
\\
\rowcolor{gray!20}
\textit{Proposed Method}
& LifeMem
& \textbf{2.6879} & \textbf{0.2519} & \textbf{0.2932} & \textbf{0.3060}
\\
\bottomrule
\end{tabular}
\caption{Overall results on Understanding Society across three models. Within-Group Gap, Entropy Gap, and Transition JS denote within-group pairwise distance gap, normalized entropy gap, and transition-distribution JS divergence. Lower values are better for all metrics. Bold marks the best result for each model and metric; LifeMem is shaded light gray. $^{*}$ indicates results significantly worse than LifeMem in a two-sided paired $t$-test conducted over shared questions ($p<0.05$).}
\label{tab:main-results}
\end{table*}

This section evaluates LifeMem's effectiveness, mechanisms, scaling, and efficiency. We examine explicit-memory limitations, overall performance, parameter differentiation, component contributions, and computational costs.

\subsection{Limits of Explicit Prompt Memory}
\label{sec:explicit-memory-bottleneck}

\begin{table*}[t]
\centering
\setlength{\tabcolsep}{4pt}
\begin{tabular}{l|cccc}
\toprule
\textbf{Method}
&
KL Divergence $\downarrow$&
Within-Group Gap $\downarrow$&
Entropy Gap $\downarrow$&
Transition JS $\downarrow$
\\
\midrule

\multicolumn{5}{c}{\textbf{\textsc{Llama-3.1-8B-Instruct}}}\\
\midrule
\rowcolor{gray!20}
LifeMem (Full)
& \textbf{3.4529}
& \textbf{0.2886}
& \textbf{0.3742}
& \textbf{0.3331}
\\
LifeMem without Parametric Memory
& 5.0874$^{*}$
& 0.3259$^{*}$
& 0.4150$^{*}$
& 0.3594$^{*}$
\\
LifeMem without Structured Memory
& 4.7555$^{*}$
& 0.3379$^{*}$
& 0.4318$^{*}$
& 0.3614$^{*}$
\\
\midrule
\multicolumn{5}{c}{\textbf{\textsc{Ministral-3-8B-Instruct-2512}}}\\
\midrule
\rowcolor{gray!20}
LifeMem (Full)
& \textbf{1.9659}
& \textbf{0.2277}
& \textbf{0.2659}
& \textbf{0.3399}
\\
LifeMem without Parametric Memory
& 3.8454$^{*}$
& 0.2810$^{*}$
& 0.3389$^{*}$
& 0.3838
\\
LifeMem without Structured Memory
& 2.4410$^{*}$
& 0.2483$^{*}$
& 0.3093$^{*}$
& 0.3443
\\
\midrule
\multicolumn{5}{c}{\textbf{\textsc{Qwen3.5-9B}}}\\
\midrule
\rowcolor{gray!20}
LifeMem (Full)
& \textbf{2.6879}
& \textbf{0.2519}
& \textbf{0.2932}
& \textbf{0.3060}
\\
LifeMem without Parametric Memory
& 3.4224$^{*}$
& 0.2738$^{*}$
& 0.3193$^{*}$
& 0.3411$^{*}$
\\
LifeMem without Structured Memory
& 3.2797$^{*}$
& 0.2842$^{*}$
& 0.3430$^{*}$
& 0.3235$^{*}$
\\
\bottomrule
\end{tabular}
\caption{Ablation study of LifeMem on Understanding Society. Within-Group Gap, Entropy Gap, and Transition JS denote within-group pairwise distance gap, normalized entropy gap, and transition-distribution JS divergence. All metrics are lower-is-better. Bold marks the best result for each model and metric; LifeMem (Full) is shaded light gray. $^{*}$ indicates results significantly worse than LifeMem (Full) in a two-sided paired $t$-test conducted over shared questions ($p<0.05$).}
\label{tab:ablation-results}
\end{table*}

\begin{figure}[t]
    \centering
    \includegraphics[width=\columnwidth]{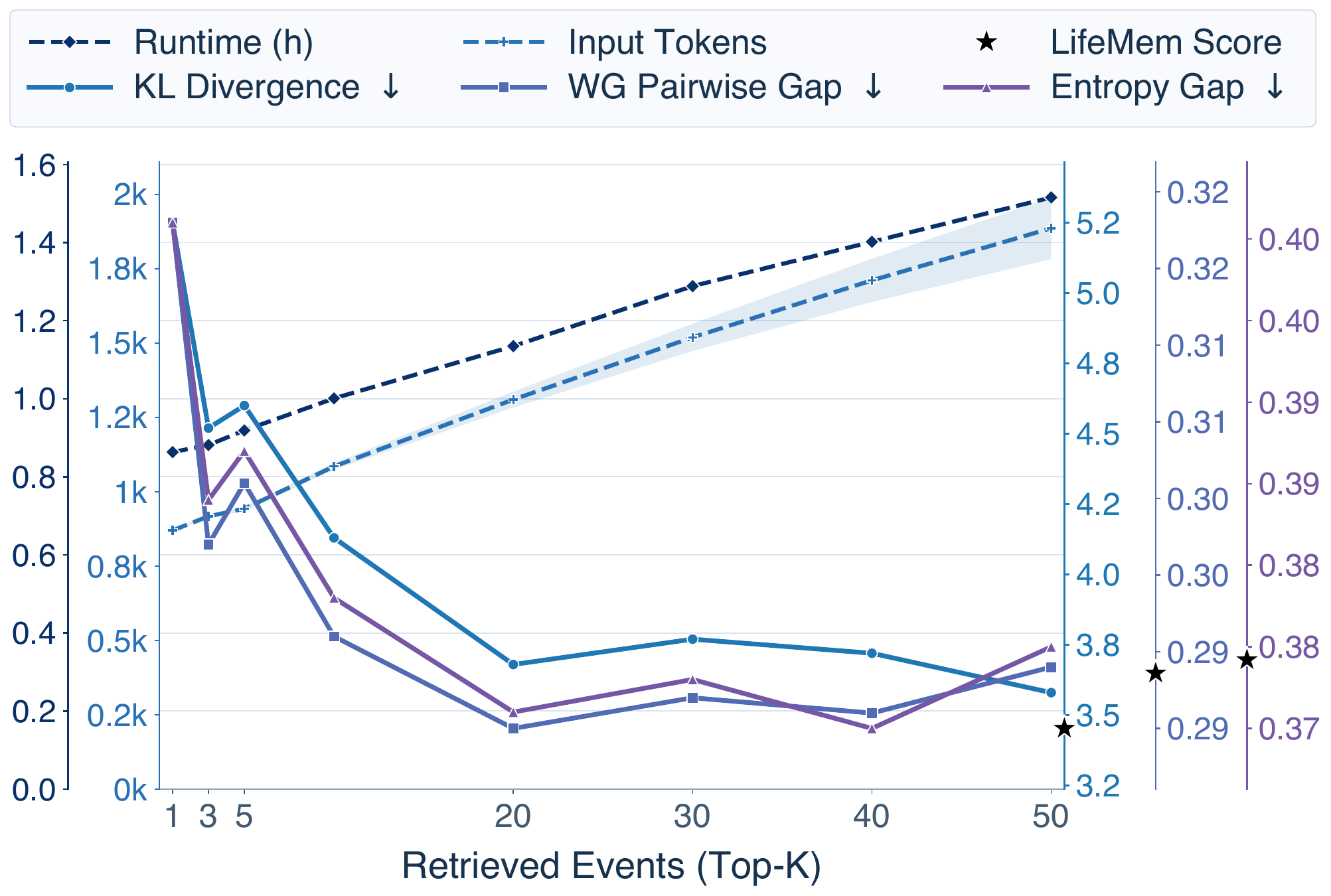}
    \caption{Effect of \textsc{Event RAG} retrieval depth on Understanding Society with Llama-8B across 100 agents. We report alignment metrics, runtime, and input tokens for top-$K\in[1,50]$. Stars mark LifeMem at $K=5$. The shaded bands indicate standard deviation across waves.}
    \label{fig:event-rag-topk}
\end{figure}

\begin{figure}[t]
    \centering
    \includegraphics[width=0.9\columnwidth]{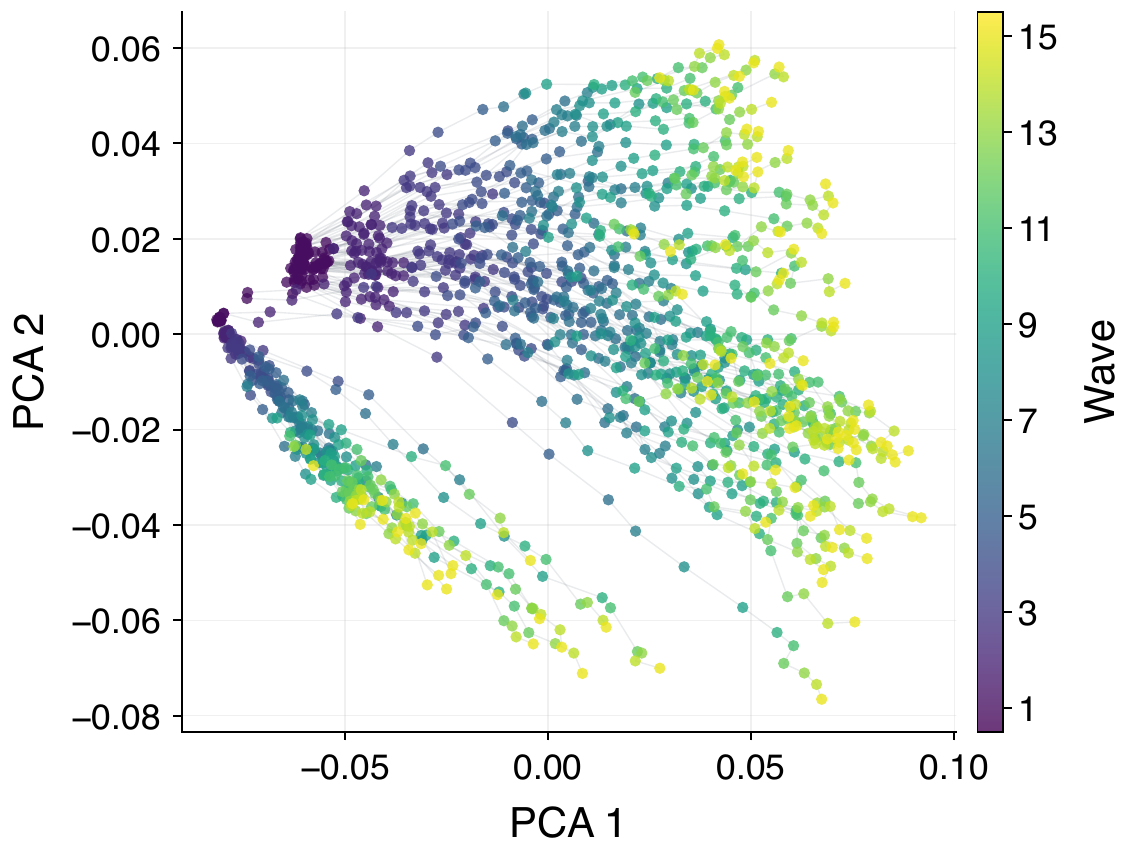}
    \caption{PCA visualization of LifeMem’s agent-specific LoRA states for 100 Understanding Society agents using Llama-8B. Points represent states after each wave, with trajectories connecting the same agent over time. Agent states increasingly diverge as life histories accumulate.}
    \label{fig:lora-us-llama}
\end{figure}

We test whether retrieving more explicit life-history information is sufficient for longitudinal simulation by varying \textsc{Event RAG} retrieval depth from $K=1$ to $K=50$ on Understanding Society with Llama-8B. We evaluate distributional and diversity alignment together with runtime and average input-token use to characterize both effectiveness and computational cost.

As shown in Figure~\ref{fig:event-rag-topk}, the three simulation metrics improve substantially at smaller retrieval depths but do not improve uniformly beyond $K=20$, while runtime and input-token use continue to increase. This trade-off illustrates a practical limitation of prompt-only memory and motivates combining selective retrieval with persistent parametric memory.

\subsection{Overall Performance and Diversity}
\label{sec:main-results}

Table~\ref{tab:main-results} shows that LifeMem performs well across four dimensions, with largely significant gains across three backbones on Understanding Society. It reduces distributional and diversity gaps relative to human responses across settings. \textsc{Direct} performs similarly poorly across models, while Ministral shows the largest improvement with LifeMem.

\textsc{Profile} improves over \textsc{Direct} but remains behind LifeMem, suggesting that static demographic profiles cannot fully capture human heterogeneity and temporal variation in these settings.
\textsc{Multilingual} and \textsc{Anti-Stereotype} prompting yield inconsistent gains, suggesting that these prompting strategies do not reliably reproduce human-like diversity across settings.
\textsc{SimVBG}, \textsc{Full History}, and \textsc{Event RAG} perform better than static baselines but still generally trail LifeMem, suggesting that non-parametric memory alone may be insufficient to achieve the same level of alignment.
LifeMem also outperforms \textsc{Random Event}, highlighting the importance of respondent-specific trajectories.

Overall, LifeMem improves alignment with human data across response distributions, overall and within-group diversity, and longitudinal response dynamics.
These results suggest that LifeMem mitigates patterns consistent with identity essentialism by reducing within-group homogenization while maintaining closer alignment with human responses.

\subsection{Ablation Study}
\label{sec:ablation}

Table~\ref{tab:ablation-results} evaluates the contributions of structured and parametric memory. Removing either component significantly degrades performance on most metrics, supporting their complementarity. Parametric memory maintains a persistent agent state that integrates experiences across waves beyond the events retrieved for the current question, whereas structured memory supplies explicit, traceable, and temporally grounded evidence for generation and tends to contribute more strongly to diversity alignment. Their relative effects vary across different models.

\subsection{Evolution of Agent-Specific LoRA States}
\label{sec:lora-visualization}

Figure~\ref{fig:lora-us-llama} uses PCA to project the high-dimensional agent-specific LoRA states into a shared two-dimensional space, illustrating how the states of 100 Llama-8B agents evolve over 15 waves on Understanding Society.
The initially concentrated adapters become progressively more dispersed and follow distinct trajectories as individual histories accumulate over time. 
This pattern suggests that sequential updates gradually encode heterogeneous life experiences into differentiated parametric states, reflecting increasing personalization across agents with different longitudinal trajectories.

\subsection{Scaling with Accumulated Life Experiences}
\label{sec:event-scaling}

\begin{figure}[t]
    \centering
    \includegraphics[width=\columnwidth]{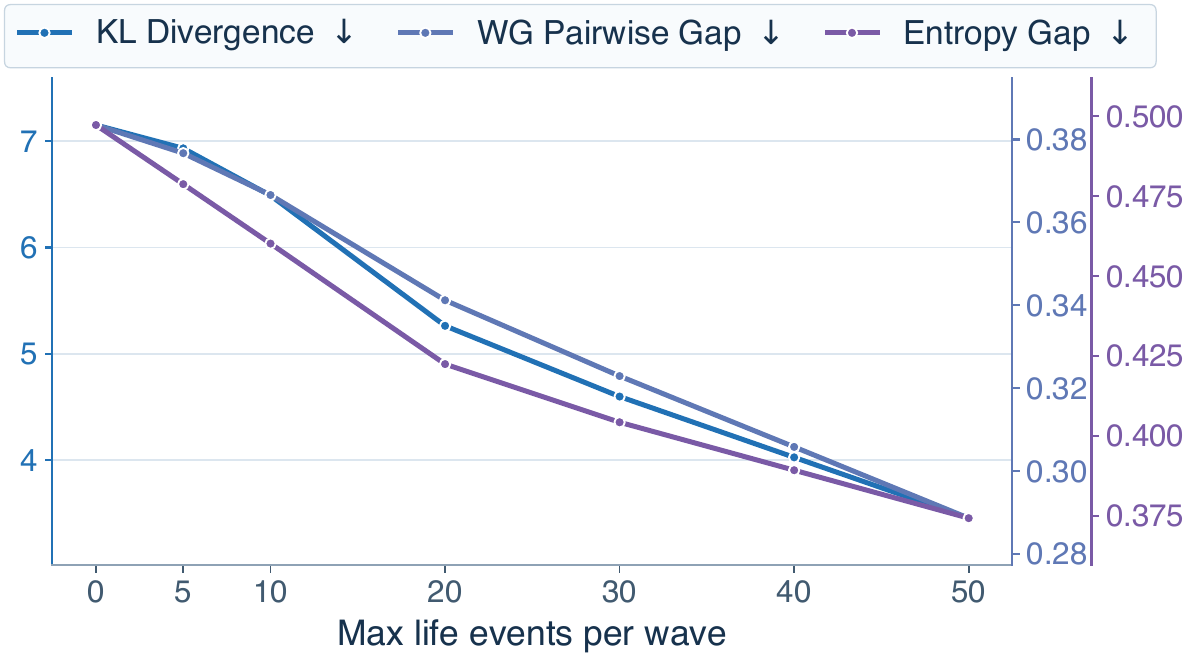}
    \caption{Effect of life-event coverage in LifeMem on Understanding Society with Llama-8B. We simulate 100 agents while varying the maximum number of life events available per wave from 0 to 50. Curves report KL divergence, within-group pairwise distance gap, and normalized entropy gap. Increasing life-event coverage improves alignment with human response distributions and diversity.}
    \label{fig:event-count-metrics}
\end{figure}

Figure~\ref{fig:event-count-metrics} examines the effect of longitudinal life-event coverage in LifeMem on Understanding Society with Llama-8B. We vary the maximum number of available events per wave from 0 to 50 for 100 agents.
Increasing coverage reduces KL divergence, normalized entropy gap, and within-group pairwise distance gap across the evaluated range.
This pattern indicates that richer longitudinal histories provide more informative and individualized evidence for distinguishing individuals and recovering more human-like response distributions and levels of diversity over time. Overall, these results further support our motivation to model agents through accumulated life experiences rather than relying on static demographic profiles alone.

\subsection{Efficiency Analysis}
\label{sec:cost-analysis}

All efficiency measurements are obtained under the same hardware configuration. We focus primarily on relative trends and comparisons across methods, as absolute runtime may vary with different hardware characteristics, including CPU, GPU, memory, and storage configurations.

\noindent\textbf{Online inference.}
Table~\ref{tab:online-inference-time} compares the per-question inference latency across methods on Understanding Society using Llama-8B.
Retrieval latency is included for both the \textsc{Event RAG} and LifeMem methods to ensure a consistent end-to-end comparison.
LifeMem is slower than lightweight prompt-only baselines but remains faster than \textsc{Full History}, \textsc{Event RAG}, and \textsc{SimVBG} on Understanding Society.
Adapter loading introduces little additional latency and can be amortized across questions once an agent's adapter is loaded.

\begin{table}[t]
\centering
\begin{tabular}{l|rr}
\toprule
\textbf{Method} & \textbf{Time} & \textbf{Rel.} \\
\midrule
Direct           & 14.4           & 1.0  \\
Profile          & 70.8           & 4.9  \\
Multilingual     & 26.5           & 1.8  \\
Anti-Stereotype  & 72.9           & 5.1  \\
SimVBG           & 196.4          & 13.7 \\
Full History     & 318.1          & 22.1 \\
Event RAG        & 956.2          & 66.5 \\
Random Event     & 79.8           & 5.6  \\
\rowcolor{gray!20}
LifeMem          & 154.5$_{+1.4}$  & 10.7 \\
\bottomrule
\end{tabular}
\caption{Average per-question inference latency with Llama-8B. Time is reported in milliseconds, and Rel. denotes latency relative to \textsc{Direct}. Retrieval latency is included for \textsc{Event RAG} and LifeMem. For LifeMem, the subscripted $+x$ indicates the LoRA adapter-loading latency. Rel. excludes the one-time LoRA adapter-loading latency.}
\label{tab:online-inference-time}
\end{table}

\begin{figure}[t]
    \centering
    \includegraphics[width=\columnwidth]{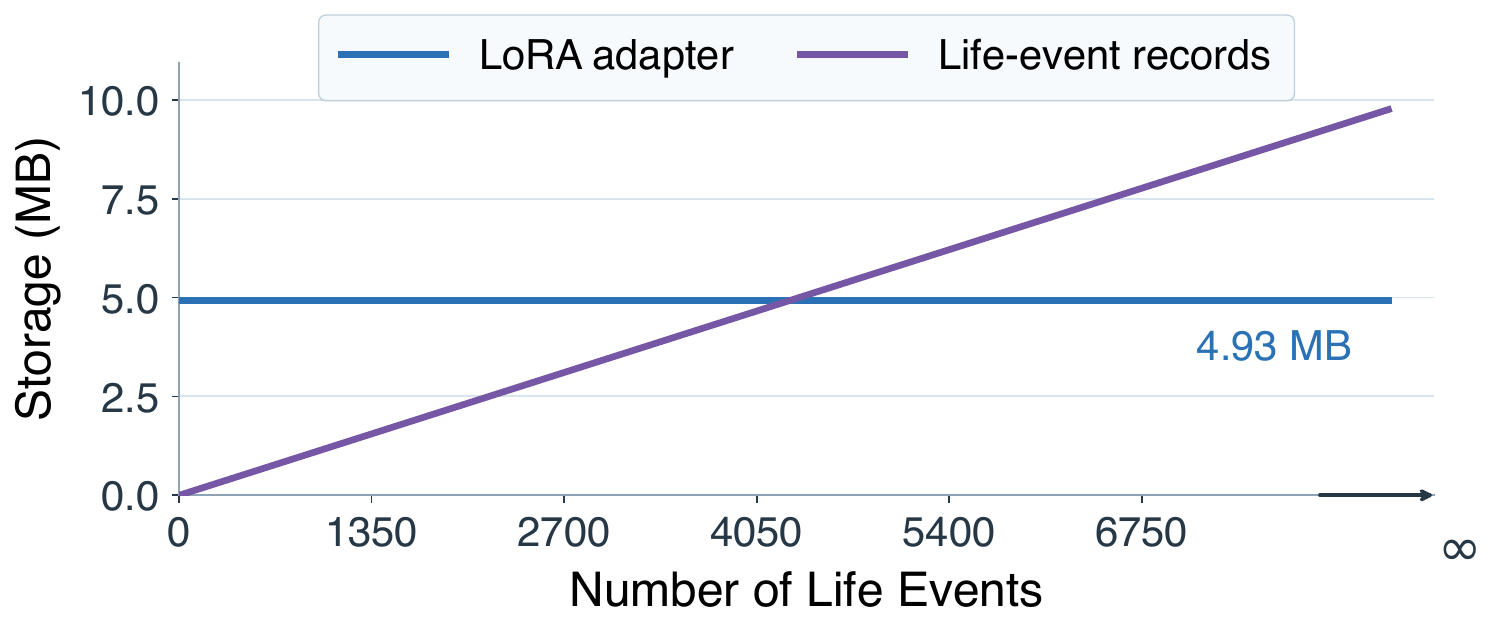}
    \caption{Per-agent storage with Llama-8B as life events accumulate. The fixed-size LoRA adapter remains constant, while structured life-event storage grows with trajectory length.}
    \label{fig:storage-life-events}
\end{figure}

\noindent\textbf{Storage.}
Figure~\ref{fig:storage-life-events} compares per-agent adapter and structured life-event storage for Llama-8B as longitudinal histories accumulate. Structured event storage grows with trajectory length because newly observed experiences must be retained explicitly, whereas each agent maintains a single fixed-size LoRA adapter that is updated after each wave. The updated adapter is then used for subsequent inference, so parametric-memory storage remains bounded rather than growing with the number of accumulated events.

\noindent\textbf{Offline training.}
With Llama-8B, the average offline adapter update time per agent and wave is 2.63 seconds on Understanding Society. This one-time cost is incurred during wave-level updates and is not repeated for each online query.

\section{Conclusion}
\label{sec:conclusion}

This work aims to mitigate patterns consistent with identity essentialism in LLM social simulation, where static profiles can suppress individual diversity and temporal change. LifeMem combines structured life-event retrieval with agent-specific LoRA parametric memory. Across Understanding Society and three model backbones, it consistently improves alignment with human response distributions, diversity, and changes across life stages. Overall, the results suggest that, in these settings, incorporating longitudinal life experiences together with persistent parametric memory provides a more faithful representation of individuals.

\section{Ethical Statement}
\label{sec:ethics}

We use only public-use, de-identified survey data, comply with applicable data-use agreements, avoid re-identification, and report aggregate results. Because agent-specific adapters may memorize respondent information, we do not release respondent-level adapters. These agents are not digital replicas and should not be used for consequential person-level decisions; they are intended for methodological research on longitudinal LLM social simulation.

\section{Acknowledgements}
This work is supported by the Research Project of Quancheng Laboratory, China (Grant No.~QCL20250105), the National Natural Science Foundation of China No.~62502260, and the Postdoctoral Fellowship Program of China Postdoctoral Science Foundation No.~GZC20240833.

\bibliography{aaai2027}

\tcbset{
    promptboxstyle/.style={
        enhanced,
        breakable,
        colback=gray!7,
        colframe=black!55,
        coltitle=white,
        colbacktitle=gray!90,
        fonttitle=\small\sffamily\bfseries,
        fontupper=\small\sffamily,
        arc=1mm,
        boxrule=0.5mm,
        top=1mm,
        bottom=1mm,
        left=2mm,
        right=2mm,
        before skip=6pt,
        after skip=6pt,
        before upper={
            \setlength{\parindent}{0pt}
            \setlength{\parskip}{0.6\baselineskip}
        }
    }
}

\newtcolorbox{promptbox}[2][]{
    promptboxstyle,
    title={#2},
    #1
}

\appendix
\section{Limitations}
\label{app:limitations}

LifeMem is limited by the coverage and granularity of longitudinal surveys, which cannot fully capture an individual's life experiences. 
Surveys may omit events outside the questionnaire, record them only at coarse intervals, and fail to capture event duration or respondents' subjective interpretations.
Thus, reconstructed trajectories remain partial representations of personal development rather than complete life histories. 
Moreover, self-reported survey responses may be affected by recall bias and question framing, and do not necessarily correspond to real-world behavior.

\section{Details of the Motivating Analyses}
\label{app:motivating-experiments}

\subsection{WVS Static-Profile Experiment}
\label{app:wvs-experiment}

\paragraph{Experimental setup.}
We use Wave~7 of the World Values Survey (WVS) \citep{haerpfer2022world} and randomly sample 2,000 respondents with seed 42. 
For each respondent, we construct a static profile from answers in the WVS demographic section, corresponding to Questions Q260--Q290. 
The profile is provided to \texttt{Llama-3.1-8B-Instruct}, which is instructed to simulate the corresponding respondent and answer the survey questions. 
This setup represents a typical profile-conditioned social simulation in which each agent is reconstructed from demographic attributes.

\paragraph{Prompt template.}
The following prompt template is instantiated with the recoded demographic attributes of each respondent:

\begin{promptbox}{Prompt Template}
Forget you are an AI model. Simulate a human being. Please answer the following question truthfully.

Please answer based on the following personal profile:

You are <gender>. You are aged <age>. You live in <country>. You are <citizenship status>. You were born <birthplace status>. Your mother was born <mother's birthplace status>. Your father was born <father's birthplace status>. You live in <urbanicity>, <settlement type>, with a population of <population size>. Your household consists of <household size> members. You <parent cohabitation status>. You speak <home language> at home. You are <marital status> and <children description>. Your education level is <education level>. Your spouse's education level is <spouse education level>. You are <employment status>. You work as <occupation>. You work in <employment sector>. Your spouse is <spouse employment status>. Your spouse works as <spouse occupation>. You are <chief wage earner status>. During the past year, your family <household financial change>. You belong to <subjective social class>. Your income is in income group <income group> (1 = lowest, 10 = highest). You identify as <religion>. Your ethnic group is <ethnic group>.

\textbf{Question:} \texttt{<question>}

\textbf{Options:} \texttt{<options>}

Please ONLY output a number. Do NOT write any words, symbols, punctuation, or explanations.
\end{promptbox}

\paragraph{Response processing.}
For both human respondents and their corresponding agents, the survey answers are converted into response vectors. Each question dimension is standardized before dimensionality reduction, after which PCA is applied to obtain two-dimensional representations. Human and agent responses follow the same response labeling, processing, standardization, PCA, and visualization procedure.

\paragraph{Socioeconomic status construction.}
We construct a composite socioeconomic status (SES) measure from four WVS variables: subjective social class (Q287), household economic situation (Q286), educational attainment (Q275), and income group (Q288).

Each variable is converted to a directionally consistent numerical scale such that larger values indicate higher socioeconomic status. 
Let $\widetilde{x}_{im}$ denote the recoded value of variable $m$ for respondent $i$, and let $\mathcal{M}_i$ denote the set of available SES variables for that respondent. The composite score is

\begin{equation}
\operatorname{SES}_i
=
\frac{1}{|\mathcal{M}_i|}
\sum_{m \in \mathcal{M}_i}
\widetilde{x}_{im}.
\end{equation}

Respondents are divided into low-, middle-, and high-SES groups according to the tertiles of the composite score.

\paragraph{Silhouette score.}
We use the silhouette score to quantify how strongly the response representations separate according to the predefined SES groups. Let $C_i$ denote the SES group containing individual $i$, and let $d(i,j)$ denote the Euclidean distance between the response representations of individuals $i$ and $j$. The average distance between $i$ and other members of the same group is

\begin{equation}
a(i)
=
\frac{1}{|C_i|-1}
\sum_{\substack{j \in C_i \\ j \neq i}}
d(i,j).
\end{equation}

The minimum average distance between $i$ and any other SES group is

\begin{equation}
b(i)
=
\min_{C \neq C_i}
\frac{1}{|C|}
\sum_{j \in C}
d(i,j).
\end{equation}

The silhouette value for individual $i$ is

\begin{equation}
s(i)
=
\frac{b(i)-a(i)}
{\max\{a(i),b(i)\}},
\end{equation}

and the overall silhouette score is

\begin{equation}
S
=
\frac{1}{N}
\sum_{i=1}^{N}
s(i).
\end{equation}

Silhouette scores are computed in the standardized response space before PCA projection; PCA is used only for visualization.

A score close to 1 indicates compact within-group representations and clear between-group separation, a score close to 0 indicates substantial overlap, and a negative score indicates that some individuals are closer to another group than to their assigned group.

In this work, the silhouette score is used only to quantify the extent to which response representations align with predefined demographic groups. It is not interpreted as a standalone measure of simulation quality.

\paragraph{Result summary.}
Human responses are broadly dispersed and substantially overlap across SES groups, resulting in a silhouette score of $S=-0.02$. In contrast, agent responses form more compact within-group clusters and show clearer separation between SES groups, resulting in $S=0.19$. This pattern suggests that static demographic prompting may compress complex individuals into group-typical representatives, thereby reducing within-group heterogeneity and amplifying between-group differences. 
We treat this experiment as a motivating diagnostic that reveals a pattern consistent with identity essentialism in LLM agents.

\subsection{Event RAG Retrieval-Depth Analysis}
\label{app:event-rag-depth}

We analyze how retrieval depth affects the effectiveness and computational cost of the \textsc{Event RAG} baseline.
Experiments are conducted on Understanding Society using Llama-3.1-8B-Instruct and 100 respondents sampled with random seed 42.
Event retrieval uses \texttt{all-MiniLM-L6-v2}, consistent with the retrieval component used by LifeMem in this comparison. Its lightweight architecture keeps retrieval overhead low while maintaining competitive retrieval quality.

We vary the retrieval depth (Top-$K$) from 1 to 50. For each value of $K$,
\textsc{Event RAG} retrieves the $K$ highest-ranked life events relevant to each evaluation question and inserts them into the model prompt. 
We report KL divergence, within-group pairwise distance gap, and normalized entropy gap, where lower values indicate closer agreement with the corresponding human distributions.

\paragraph{Runtime.}
Runtime is measured as the end-to-end wall-clock duration of one complete \textsc{Event RAG} experiment over all 100 respondents and all fifteen survey waves. It therefore includes model loading, population initialization, life-event loading, event retrieval, prompt construction, model inference, output parsing, and metric computation.

\paragraph{Input tokens.}
For each retrieval depth $K$, we average the estimated input tokens over all respondent--question prompts across all fifteen survey waves. The reported value therefore represents the mean input length per evaluation prompt.

\paragraph{Comparison with LifeMem.}
Figure~\ref{fig:event-rag-topk} in the main paper additionally marks LifeMem at $K=5$ as a reference.
While increasing \textsc{Event RAG} retrieval depth initially improves distributional and diversity alignment, the gains do not remain uniform beyond $K=20$, whereas runtime and input-token use continue to increase.
LifeMem achieves better alignment than \textsc{Event RAG} at the same $K=5$ and remains competitive with deeper retrieval, suggesting that its gains are unlikely to be explained solely by retrieving more life events.
Instead, these results motivate combining selective retrieval with persistent parametric memory.

\section{Dataset Construction and Preprocessing}
\label{app:dataset-construction}

\subsection{Dataset Overviews}
\label{app:dataset-overviews}

\paragraph{Understanding Society.}
Understanding Society, also known as the UK Household Longitudinal Study (UKHLS), is an annual household panel survey that began in 2009 \citep{buck2012understanding}. We use UKHLS Waves~1--15, covering interviews conducted from 2009 to 2024. Each wave is fielded over an overlapping period, while individual sample members are generally interviewed approximately one year apart.

Understanding Society follows members of sampled UK households and covers household composition, family relationships, education, employment, income, housing, health and well-being, finances, political attitudes, and social participation. 
Its annual panel structure and broad question coverage across fifteen waves support the evaluation of changes in respondents' answers between adjacent waves.

The public-use files contain between 41,601 and 77,495 respondents per wave across Waves~1--15. Detailed wave-level sample sizes are reported in Table~\ref{tab:usoc-wave-sizes}. Respondents are linked across waves using the stable person identifier \texttt{pidp}.

\begin{table}[t]
\centering
\begin{tabular}{cc|cc}
\toprule
\textbf{Wave} & \textbf{Respondents} &
\textbf{Wave} & \textbf{Respondents} \\
\midrule
1  & 77,308 & 9  & 52,694 \\
2  & 77,495 & 10 & 50,113 \\
3  & 70,671 & 11 & 46,965 \\
4  & 65,626 & 12 & 43,577 \\
5  & 61,512 & 13 & 41,601 \\
6  & 64,712 & 14 & 55,400 \\
7  & 59,941 & 15 & 48,896 \\
8  & 56,608 &      &        \\
\bottomrule
\end{tabular}
\caption{Number of respondents in the public-use Understanding Society files across Waves~1--15.}
\label{tab:usoc-wave-sizes}
\end{table}

\subsection{Respondent Alignment and Sampling}
\label{app:respondent-alignment}

To construct complete longitudinal trajectories, we align respondents using stable cross-wave identifiers and retain only individuals observed in every wave of the corresponding dataset. After alignment, 14,104 respondents are observed with the same \texttt{pidp} across all fifteen waves.

For the main experiments, we randomly sample 100 respondents from the aligned pool using seed 42. The sampled respondent identifiers are fixed and shared across all methods and backbone models. Consequently, all comparisons use the same individuals, survey questions, and observed life histories, thereby controlling for differences in respondent composition. Additional experiments using seeds 43 and 44 evaluate robustness to respondent sampling.

\subsection{Variable Classification}
\label{app:variable-classification}

\paragraph{Structured Question Records.}
Each selected variable is represented as a structured question record with the following fields:
\begin{itemize}
    \item Variable: the survey variable name.
    \item Section: the questionnaire section.
    \item Question: the natural-language survey question.
    \item Options: the response options for closed-ended questions.
    \item Note: additional constraints for open-ended responses.
\end{itemize}

\paragraph{Functional Categories.}
We organize survey variables into three functional categories:
\textit{demographic}, \textit{life event}, and \textit{evaluate}.
The \textit{demographic} variables are used to construct the initial profile of an agent, including background and profile attributes such as age, gender, education, employment, household composition, region, religion, and family background.
The \textit{life event} variables describe longitudinal experiences and state changes, such as schooling, work transitions, partnership changes, fertility, housing changes, health events, caregiving, economic hardship, and other major events. These variables are used to construct structured memory and parametric memory updates across waves. 
The \textit{evaluate} variables are reserved exclusively for model evaluation and are excluded from profile construction, structured memory, and parametric-memory updates. Examples include questions about residential preferences, internet use, social participation, political attitudes, and subjective well-being.

\begin{table}[t]
\centering
\begin{tabular}{llrrr}
\toprule
Dataset & Wave & Demo. & Event & Eval. \\
\midrule
\multirow{15}{*}{UKHLS}
& 1  & 68 & 72 & 11 \\
& 2  & 62 & 37 & 11 \\
& 3  & 67 & 55 & 38 \\
& 4  & 69 & 55 & 37 \\
& 5  & 61 & 69 & 49 \\
& 6  & 66 & 85 & 41 \\
& 7  & 62 & 54 & 26 \\
& 8  & 69 & 47 & 16 \\
& 9  & 68 & 68 & 34 \\
& 10 & 68 & 44 & 28 \\
& 11 & 71 & 56 & 50 \\
& 12 & 65 & 57 & 64 \\
& 13 & 70 & 65 & 25 \\
& 14 & 70 & 66 & 43 \\
& 15 & 65 & 89 & 45 \\
\bottomrule
\end{tabular}
\caption{Numbers of demographic, life-event, and evaluation variables selected for each survey wave. Demo., Event, and Eval. denote demographic, life-event, and evaluation variables, respectively. UKHLS denotes Understanding Society.}
\label{tab:variable-counts}
\end{table}

\paragraph{Classification and Filtering.}
Variables are classified using a rule-based semantic matching procedure followed by response-quality filtering. 
The semantic classifier considers the section, question, options, note, and variable fields.
We maintain category-specific keyword dictionaries for the three categories. After semantic matching, we filter variables using response-quality statistics computed from the human answer files. 
For each variable, we compute the total number of responses, the number of valid responses, and the number and rate of missing responses. Variables with severe missingness or too few valid responses are removed. Table~\ref{tab:variable-counts} reports the number of selected variables in each category for every wave.

\paragraph{Examples of Classified Variables.}
The following examples illustrate the three variable categories using questions from Understanding Society Wave~1.

\begin{promptbox}{Demographic Variables}

\textbf{Variable:} \texttt{Marstat}

\textbf{Section:} Household Grid

\textbf{Question:} What is [person]'s legal marital status

\textbf{Options:}\\
1: Single and never married or never in a legally recognised civil partnership;\\
2: Married;\\
3: In a registered same-sex civil partnership;\\
4: Separated but legally married;\\
5: Divorced;\\
6: Widowed;\\
7: Separated from civil partner;\\
8: A former civil partner, the civil partnership legally dissolved;\\
9: A surviving civil partner (your partner having died)

\medskip
\hrule
\medskip

\textbf{Variable:} \texttt{Livewith}

\textbf{Section:} Household Grid

\textbf{Question:} May I just check, are you/is living with someone in this household as a couple?

\textbf{Options:}\\
1: Yes;\\
2: No;\\
3: Same sex couple

\end{promptbox}

\begin{promptbox}{Life-Event Variables}

\textbf{Variable:} \texttt{Matdepb}

\textbf{Section:} Household Questionnaire

\textbf{Question:} Friends or family around for a drink or meal at least once a month?

\textbf{Options:}\\
1: I/We have this;\\
2: I/We would like to have this but cannot afford this at the moment;\\
3: I/We do not want/need this at the moment

\medskip
\hrule
\medskip

\textbf{Variable:} \texttt{Cplay}

\textbf{Section:} Household Questionnaire

\textbf{Question:} Does your child/do your children go to a toddler group, nursery or playgroup at least once a week?

\textbf{Options:}\\
1: Child(ren) do this;\\
2: Child(ren) would like to do this but we cannot afford this at the moment;\\
3: Child(ren) do not want/need this at the moment

\end{promptbox}

\begin{promptbox}{Evaluation Variables}

\textbf{Variable:} \texttt{Lkmove}

\textbf{Section:} Demographics

\textbf{Question:} If you could choose, would you stay here in your present home or would you prefer to move somewhere else?

\textbf{Options:}\\
1: Stay here;\\
2: Prefer to move

\medskip
\hrule
\medskip

\textbf{Variable:} \texttt{Vote2}

\textbf{Section:} Politics

\textbf{Question:} Do you think of yourself as a little closer to one political party than to the others?

\textbf{Options:}\\
1: Yes;\\
2: No

\end{promptbox}

\subsection{Common Longitudinal Evaluation Set}
\label{app:common-longitudinal-eval}

In addition to the wave-specific evaluation variables, we construct a common longitudinal evaluation set for Understanding Society. This set provides a fixed target space across waves, enabling us to evaluate whether agents reproduce not only population-level response distributions but also changes in the same survey variables over time.

We first retain variables that appear in the question files for all 15 waves, yielding 198 shared variables. We then exclude variables used as demographic or life-event inputs, leaving 136 candidates. After semantic and response-quality filtering, the final evaluation set contains 50 variables covering housing, employment, income, caregiving, family, education, and related life outcomes.

Transition-distribution JS divergence is computed on Understanding Society using a fixed set of comparable evaluation variables across all 15 waves.

\section{Shared Experimental Setup}
\label{app:shared-setup}

\subsection{Inference Protocol}
\label{app:inference-protocol}

We evaluate all methods using three instruction-tuned backbone models:
Llama-3.1-8B-Instruct, Ministral-3-8B-Instruct-2512, and Qwen3.5-9B.
Thinking mode is disabled for Qwen3.5-9B, and each model uses its native tokenizer and chat template. 
All methods share the same deterministic decoding configuration, with temperature $0$, sampling disabled, a maximum input length of 4,096 tokens, and a maximum output length of 16 tokens. Inference is conducted with a batch size of four using \texttt{bfloat16} precision.

All methods also share the same survey-response prompt structure. Method-specific context, such as a demographic profile or retrieved life events, is inserted before the question when applicable. For closed-ended questions, the model is required to return exactly one option number. For open-ended questions, it is instead instructed to provide a concise response without explanation.

\begin{promptbox}{Shared Survey Prompt}
You are not an AI assistant; you are role-playing a human survey respondent.

\textbf{Question:} \texttt{<question>}

\textit{For closed-ended questions:} 

\textbf{Options:} \texttt{<options>}

Answer with exactly one option number. Do not provide any explanation.

\textit{For open-ended questions:} 

\textbf{Notes:} \texttt{<notes>}

Answer concisely. Do not provide any explanation.
\end{promptbox}

\subsection{Survey-to-Statement Conversion}
\label{app:survey-to-statement}

We convert each survey question--answer pair into one second-person statement and one first-person statement.
The second-person statements are used to construct demographic profiles and structured life-event memory, while the first-person statements serve as reconstruction targets for parametric-memory training.

Conversion is performed using gpt-3.5-turbo-0125, with temperature $0$ and a maximum output length of 2,048 tokens.
The conversion prompt is shown below.

\begin{promptbox}{Survey-to-Statement Conversion Prompt}
You convert survey question-answer pairs into concise personal statements.

Rewrite the provided question and answer as one faithful second-person statement and one first-person statement about the respondent.

\textbf{Rules:}

1. Preserve the exact meaning of the answer.

2. Do not add information, explanations, causes, stereotypes, or assumptions.

3. Preserve negation, quantities, dates, frequencies, and time ranges.

4. For life events, do not convert an event into a permanent trait unless the answer says so.

5. Return valid JSON only with keys \texttt{second\_person} and \texttt{first\_person}.

\textbf{Input:} \texttt{<input>}
\end{promptbox}

For example, the survey response

\begin{promptbox}{Example Survey Response}
\textbf{Question:} How often have you had trouble falling asleep or staying asleep?

\textbf{Answer:} Never
\end{promptbox}

is converted into:

\begin{promptbox}{Example Converted Statements}
\textbf{Second\_Person:} You have never had trouble falling asleep or staying asleep.

\textbf{First\_Person:} I have never had trouble falling asleep or staying asleep.
\end{promptbox}

This conversion preserves the observed survey response while producing textual representations suitable for profile construction, event retrieval, and adapter training.

\section{Baseline Implementation Details}
\label{app:baseline-implementations}

\subsection{Static Conditioning}

\paragraph{\textsc{Direct}.}
\textsc{Direct} \citep{argyle2023out,santurkar2023whose,bisbee2024synthetic,hu2024quantifying} receives only the shared survey instruction, the current question, its response options or notes, and the corresponding output constraint.
It does not receive demographic information or longitudinal life events.

\paragraph{\textsc{Profile}.}
\textsc{Profile} \citep{argyle2023out,santurkar2023whose,bisbee2024synthetic,hu2024quantifying} additionally conditions the model on each respondent's fixed initial demographic profile.

\begin{promptbox}{Profile Conditioning Prompt}
You are answering a longitudinal social survey as the described person.

\textbf{Demographic profile:} \texttt{<profile\_text>}
\end{promptbox}

\subsection{Diversity-Oriented Prompting}

\paragraph{\textsc{Multilingual}.}
\textsc{Multilingual} \citep{wang2025multilingual} follows the \textsc{Direct} setting but translates the complete survey prompt into four widely used languages: Chinese, Spanish, English, and Arabic. For each respondent-question pair, one of the four languages is selected uniformly at random using a deterministic hash of the global seed, wave index, respondent ID, and question variable.

\paragraph{\textsc{Anti-Stereotype}.}
\textsc{Anti-Stereotype} \citep{sivakumar2025bias} follows the \textsc{Profile} setting and appends an explicit instruction discouraging demographic stereotyping. This baseline evaluates whether explicit anti-stereotyping guidance can mitigate demographic overgeneralization.

\begin{promptbox}{Anti-Stereotype Instruction}
Do not assume that one demographic attribute determines another. When information is missing, preserve uncertainty rather than filling it with stereotypes.
\end{promptbox}

\subsection{Non-Parametric Memory}

\paragraph{\textsc{SimVBG}.}
\textsc{SimVBG} \citep{du2025simvbg} first uses the current backbone model to expand each respondent's demographic profile into a comprehensive second-person background story, which remains fixed across all survey waves.
Story generation uses a batch size of four, temperature $0$, sampling disabled, and a maximum output length of 2,048 tokens.

The generation prompt includes the following instructions:

\begin{promptbox}{Background-Story Generation Prompt}
You are a background story writer. Your task is to craft a comprehensive backstory for a person based on the information provided below. You must include every single data point from the original information, without exception.

\textbf{This person's information:} \texttt{<profile\_text>}
\end{promptbox}

During evaluation, the generated story is inserted before the shared survey prompt:

\begin{promptbox}{SimVBG Evaluation Prompt}
You are answering a longitudinal social survey as the person described in the background story.

\textbf{Background story:} \texttt{<background\_story>}
\end{promptbox}

\paragraph{\textsc{Full History}.}
\textsc{Full History} \citep{maharana2024evaluating} conditions the model on the demographic profile and all respondent-specific life events accumulated through the current wave:

\begin{promptbox}{Full-History Prompt}
You are answering a longitudinal social survey as the described person.

\textbf{Demographic profile:} \texttt{<profile\_text>}

\textbf{Life history:}

-- \texttt{<event\_1>}

-- \texttt{<event\_2>}

...
\end{promptbox}

When the accumulated history exceeds the available input budget, events are retained in reverse chronological order under a character budget of
$3\times\texttt{max\_input\_length}$.
Recent events are therefore preserved first, while older events may be truncated.

\paragraph{\textsc{Event RAG}.}
\textsc{Event RAG} \citep{lewis2020rag,maharana2024evaluating}
conditions the model on the demographic profile and the five accumulated life events most relevant to the current evaluation question.
It uses \texttt{bge-m3} to encode each evaluation question and each life-event text, where a life-event text concatenates the wave index, survey section, original question, answer text, and second-person statement.

For an event $e_{i,\tau}$ observed for respondent $i$ at wave $\tau$ and a question $q_t$ asked at wave $t$, the retrieval score is
\begin{equation}
s(q_t,e_{i,\tau})
=
\operatorname{sim}(q_t,e_{i,\tau})
\cdot
\alpha_{\mathrm{ret}}^{\,t-\tau},
\end{equation}
where $\alpha_{\mathrm{ret}}=0.9$ controls recency decay. The semantic similarity is computed as
\begin{equation}
\operatorname{sim}(q_t,e_{i,\tau})
=
\mathbf{h}_{q_t}^{\top}\mathbf{h}_{e_{i,\tau}},
\end{equation}
where $\mathbf{h}_{q_t}$ and $\mathbf{h}_{e_{i,\tau}}$ are $\ell_2$-normalized embeddings from \texttt{bge-m3}; hence the dot product is cosine similarity.
The five highest-scoring events are inserted into the prompt.

\begin{promptbox}{Event RAG Prompt}
You are answering a longitudinal social survey as the described person.

\textbf{Demographic profile:} \texttt{<profile\_text>}

\textbf{Relevant life experiences:}

-- \texttt{<retrieved\_event\_1>}

-- \texttt{<retrieved\_event\_2>}

...
\end{promptbox}

\subsection{Control Baseline}

\paragraph{\textsc{Random Event}.}
\textsc{Random Event} uses the same prompt structure and the same number of events as \textsc{Event RAG}.
Instead of retrieving respondent-specific events, it samples five events from other agents using a pseudorandom generator initialized with the global simulation seed.
This baseline preserves the amount and format of event information while removing the correspondence between the supplied events and the target respondent.

\section{LifeMem Implementation Details}
\label{app:lifemem-implementation}

LifeMem combines demographic profile conditioning, structured life-event memory, and agent-specific parametric memory.
At each survey wave, the system updates and evaluates every agent sequentially.

For each agent at every wave, LifeMem first loads the newly observed life events and updates the agent-specific LoRA adapter using both current and replayed events. It then adds the current events to structured memory, retrieves relevant events for each evaluation question, and constructs the corresponding prompt. After activating the agent-specific adapter, the model generates responses in batches, which are subsequently parsed and evaluated. Finally, the updated LoRA state is saved for use in the next wave.

\subsection{Structured Life-Event Memory}
\label{app:structured-memory-implementation}

LifeMem uses \texttt{all-MiniLM-L6-v2} as a frozen retrieval encoder.
Its lightweight architecture provides a practical balance between semantic retrieval quality and computational efficiency.
Apart from the encoder choice, its retrieval procedure, including top-$K$, similarity scoring, and recency weighting, is identical to that of \textsc{Event RAG}. The recency term favors recent experiences while allowing older events to be retrieved when their semantic relevance remains high.

\subsection{Agent-Specific Parametric Memory}
\label{app:parametric-memory-implementation}

Each agent maintains an independent LoRA adapter, while the parameters of the underlying language model remain frozen.
Table~\ref{tab:lora-implementation-settings} reports the default adapter and optimization settings.

\begin{table}[t]
\centering
\begin{tabular}{l|l}
\toprule
\textbf{Setting} & \textbf{Value} \\
\midrule
LoRA rank & 8 \\
LoRA scaling coefficient & 16 \\
LoRA dropout & 0 \\
Target modules & \texttt{q\_proj}, \texttt{v\_proj} \\
               & \texttt{o\_proj}, \texttt{down\_proj} \\
Learning rate & $1\times10^{-4}$ \\
Training epochs per update & 2 \\
Training batch size & 16 \\
Gradient accumulation steps & 1 \\
Maximum gradient norm & 1.0 \\
\bottomrule
\end{tabular}
\caption{Default configuration for agent-specific LoRA updates.}
\label{tab:lora-implementation-settings}
\end{table}

\begin{algorithm}[t]
\caption{Agent-specific parametric-memory update}
\label{alg:agent-lora-update}
\begin{algorithmic}[1]
\FOR{each wave $t$}
    \FOR{each agent $i$}
        \STATE $\mathcal{E}_{i,t}^{\mathrm{cur}}
        \leftarrow$ events of agent $i$ observed at wave $t$
        
        \STATE $\mathcal{E}_{i,t}^{\mathrm{rep}}
        \leftarrow$ sample $R$ events from agent $i$'s previous events
        
        \STATE $\mathcal{B}_{i,t} \leftarrow \emptyset$
        
        \FOR{each event $e \in \mathcal{E}_{i,t}^{\mathrm{cur}}$}
            \STATE Add $(e.\mathrm{question},e.\mathrm{answer})$
            to $\mathcal{B}_{i,t}$
            
            \STATE Add
            $(e.\mathrm{statement},e.\mathrm{first\_person})$
            to $\mathcal{B}_{i,t}$
        \ENDFOR
        
        \FOR{each event $e \in \mathcal{E}_{i,t}^{\mathrm{rep}}$}
            \STATE Add $(e.\mathrm{question},e.\mathrm{answer})$
            to $\mathcal{B}_{i,t}$
            
            \STATE Add
            $(e.\mathrm{statement},e.\mathrm{first\_person})$
            to $\mathcal{B}_{i,t}$
            \STATE Mark both examples with $\mathrm{replay}=\mathrm{true}$
        \ENDFOR
        
        \STATE Activate agent $i$'s LoRA adapter
        
        \STATE Save the pre-update parameters
        $\Theta_{i,t}^{\mathrm{before}}
        \leftarrow \Theta_{i,t}$
        
        \FOR{each training epoch}
            \FOR{each minibatch
            $\mathcal{M}\subset\mathcal{B}_{i,t}$}
                \FOR{each example $z\in\mathcal{M}$}
                    \STATE $\ell_z
                    \leftarrow
                    \operatorname{CE}\!\left(
                    f_{\Theta_{i,t}}(z_{\mathrm{input}}),
                    z_{\mathrm{assistant}}
                    \right)$
                    
                    \IF{$z$ is a replay example}
                        \STATE $\ell_z \leftarrow \eta\ell_z$
                    \ENDIF
                \ENDFOR
                
                \STATE $\mathcal{L}_{\mathrm{CE}}
                \leftarrow
                \frac{1}{|\mathcal{M}|}
                \sum_{z\in\mathcal{M}}\ell_z$
                
                \STATE $\mathcal{L}
                \leftarrow
                \mathcal{L}_{\mathrm{CE}}
                +
                \gamma
                \left\|
                \Theta_{i,t}
                -
                \Theta_{i,t}^{\mathrm{before}}
                \right\|_2^2$
                
                \STATE Backpropagate $\mathcal{L}$
                
                \STATE Clip gradients using
                \texttt{max\_grad\_norm}
                
                \STATE Update the active adapter using the optimizer
            \ENDFOR
        \ENDFOR
        
        \STATE Save agent $i$'s updated LoRA adapter state
    \ENDFOR
\ENDFOR
\end{algorithmic}
\end{algorithm}

\paragraph{Training examples.}
Each life event is converted into two training examples.
The first teaches the association between the original survey question and the respondent's observed answer:

\begin{promptbox}{Question--Answer Training Example}
\textbf{User:} <event\_question>

\textbf{Assistant:} <event\_answer\_text>
\end{promptbox}

The second reconstructs a first-person self-memory from the corresponding second-person event statement:

\begin{promptbox}{Self-Memory Reconstruction Example}
\textbf{User:} Recall the following personal information: <second\_person\_event>

\textbf{Assistant:} <first\_person\_event>
\end{promptbox}

The adapter therefore learns both the mapping from survey questions to observed responses and the reconstruction of life events as first-person personal memories.

\paragraph{Sequential adapter updates.}
At each wave, LifeMem updates every agent's adapter using the events newly observed at that wave together with replayed events sampled from the same agent's history. Algorithm~\ref{alg:agent-lora-update} summarizes the training procedure.

Replay events are sampled exclusively from the same agent's previously observed life events. By default, we uniformly sample up to $R=4$ historical events and assign replay examples a loss weight of $\eta=0.5$. Each replayed event yields the same question--answer and self-memory reconstruction examples as a current event. Sampling uses a pseudorandom generator initialized with the global simulation seed. When fewer than $R$ historical events are available, all available events are included.

Stability regularization is enabled by default. Before updating an agent at wave $t$, we store the current adapter parameters $\Theta_{i,t}^{\mathrm{before}}$ and add

\begin{equation}
\gamma
\left\|
\Theta_{i,t}
-
\Theta_{i,t}^{\mathrm{before}}
\right\|_2^2
\end{equation}

to the training objective, where $\gamma=0.01$. This term limits abrupt parameter changes and reduces the extent to which newly observed events overwrite previously accumulated parametric memory.

During inference, multiple evaluation questions for the same agent are generated in batches while that agent's LoRA adapter remains active. This avoids repeatedly switching adapters between individual questions and ensures that all responses for an agent at a given wave use the same parametric state.

\section{Evaluation Metrics}
\label{app:evaluation-metrics}

All metrics are computed only from valid human--model response pairs. Specifically, a sample is retained only when the human response corresponds to a predefined valid option and the model output can be parsed into a valid option from the same response space. 
Human responses coded as missing, refusal, ``do not know,'' or ``not applicable'' are excluded. This shared filtering rule prevents differences in missing-response behavior from affecting comparisons between methods.

Let $\mathcal{C}$ denote the set of eligible wave--question cells. For a cell $(t,q)\in\mathcal{C}$, let $\mathcal{Y}_{tq}$ be its set of valid response labels, and let $n_h^{tq}(y)$ and $n_m^{tq}(y)$ denote the numbers of valid human and model responses assigned to label $y$, respectively.

\subsection{KL Divergence}
\label{app:metric-kl}

For each wave--question cell, we construct empirical human and model response distributions. To avoid zero probabilities, we add $\epsilon=10^{-9}$ to the count of every valid label:
\begin{equation}
P_h^{tq}(y)
=
\frac{
n_h^{tq}(y)+\epsilon
}{
\sum_{y'\in\mathcal{Y}_{tq}}n_h^{tq}(y')
+
|\mathcal{Y}_{tq}|\epsilon
},
\end{equation}
and
\begin{equation}
P_m^{tq}(y)
=
\frac{
n_m^{tq}(y)+\epsilon
}{
\sum_{y'\in\mathcal{Y}_{tq}}n_m^{tq}(y')
+
|\mathcal{Y}_{tq}|\epsilon
}.
\end{equation}

We then compute the KL divergence from the human distribution to the model distribution:
\begin{equation}
\operatorname{KL}_{tq}
=
\operatorname{KL}
\left(
P_h^{tq}\,\|\,P_m^{tq}
\right)
=
\sum_{y\in\mathcal{Y}_{tq}}
P_h^{tq}(y)
\log
\frac{
P_h^{tq}(y)
}{
P_m^{tq}(y)
}.
\end{equation}

The reported KL divergence is the unweighted mean across all eligible wave--question cells:
\begin{equation}
\operatorname{KL}
=
\frac{1}{|\mathcal{C}|}
\sum_{(t,q)\in\mathcal{C}}
\operatorname{KL}_{tq}.
\end{equation}

Lower values indicate closer alignment between the model and human response distributions.

\subsection{Normalized Entropy Gap}
\label{app:metric-entropy}

For a response distribution $P$ over the valid label set $\mathcal{Y}_{tq}$, normalized entropy is defined as
\begin{equation}
H(P)
=
-
\frac{1}{\log|\mathcal{Y}_{tq}|}
\sum_{y\in\mathcal{Y}_{tq}}
P(y)\log P(y).
\end{equation}
The normalization bounds entropy between zero and one for questions with at least two valid response options. We compute the human and model entropies separately:
\begin{equation}
H_h^{tq}
=
H\!\left(P_h^{tq}\right),
\qquad
H_m^{tq}
=
H\!\left(P_m^{tq}\right).
\end{equation}

The entropy gap for each wave--question cell is the absolute difference between them:
\begin{equation}
\operatorname{EntropyGap}_{tq}
=
\left|
H_m^{tq}
-
H_h^{tq}
\right|.
\end{equation}

The reported normalized entropy gap is
\begin{equation}
\operatorname{NormEntropyGap}
=
\frac{1}{|\mathcal{C}|}
\sum_{(t,q)\in\mathcal{C}}
\operatorname{EntropyGap}_{tq}.
\end{equation}

A smaller value indicates that the overall concentration or dispersion of model responses is closer to that of human responses.

\subsection{Within-Group Pairwise Distance Gap}
\label{app:metric-wg-gap}

\begin{table}[t]
\centering
\begin{tabular}{l|l}
\toprule
\textbf{Dataset} & \textbf{Grouping variables} \\
\midrule
UKHLS &
\begin{tabular}[t]{@{}l@{}}
\texttt{Sex}, \texttt{Birthy}, \texttt{Marstat}, \texttt{Employ},\\
\texttt{Oprlg1}, \texttt{Qfhigh}, \texttt{Jbstat}
\end{tabular} \\
\bottomrule
\end{tabular}
\caption{Demographic variables used to define groups for the within-group pairwise distance metric. UKHLS denotes Understanding Society.}
\label{tab:wg-group-variables}
\end{table}

The within-group pairwise distance measures whether response diversity within demographic groups is preserved. We form groups separately using each discrete demographic variable. The grouping variables are shown in Table~\ref{tab:wg-group-variables}.

Consider demographic variable $g$, group value $v$, wave $t$, and question $q$. Let $n_h^{tqgv}(y)$ be the number of human respondents in the group who select option $y$, and let
\begin{equation}
n_h^{tqgv}
=
\sum_{y\in\mathcal{Y}_{tq}}
n_h^{tqgv}(y).
\end{equation}
For groups containing at least two valid responses, the human within-group categorical pairwise distance is
\begin{equation}
D_h(t,q,g,v)
=
1-
\frac{
\sum_{y\in\mathcal{Y}_{tq}}
n_h^{tqgv}(y)
\left(
n_h^{tqgv}(y)-1
\right)
}{
n_h^{tqgv}
\left(
n_h^{tqgv}-1
\right)
}.
\end{equation}

This quantity is the probability that two distinct respondents sampled from the same group provide different answers. The model distance is computed analogously:
\begin{equation}
D_m(t,q,g,v)
=
1-
\frac{
\sum_{y\in\mathcal{Y}_{tq}}
n_m^{tqgv}(y)
\left(
n_m^{tqgv}(y)-1
\right)
}{
n_m^{tqgv}
\left(
n_m^{tqgv}-1
\right)
}.
\end{equation}

Group cells with fewer than two valid responses are excluded. Let
$\mathcal{G}_{tq}$ denote the set of eligible demographic-variable--value cells for wave $t$ and question $q$. We first average the group-specific distances:
\begin{equation}
D_h(t,q)
=
\frac{1}{|\mathcal{G}_{tq}|}
\sum_{(g,v)\in\mathcal{G}_{tq}}
D_h(t,q,g,v),
\end{equation}
\begin{equation}
D_m(t,q)
=
\frac{1}{|\mathcal{G}_{tq}|}
\sum_{(g,v)\in\mathcal{G}_{tq}}
D_m(t,q,g,v).
\end{equation}

The wave--question-level gap is then
\begin{equation}
\operatorname{WGGap}_{tq}
=
\left|
D_m(t,q)-D_h(t,q)
\right|.
\end{equation}

Finally, the reported within-group pairwise distance gap is
\begin{equation}
\operatorname{WGPairwiseDistGap}
=
\frac{1}{|\mathcal{C}_{\mathrm{WG}}|}
\sum_{(t,q)\in\mathcal{C}_{\mathrm{WG}}}
\operatorname{WGGap}_{tq},
\end{equation}
where $\mathcal{C}_{\mathrm{WG}}$ contains wave--question cells with at least one eligible group cell. Lower values indicate closer alignment between model and human within-group response diversity.

\subsection{Transition-Distribution JS Divergence}
\label{app:metric-transition-js}

Transition-distribution JS divergence evaluates whether agents reproduce longitudinal patterns of response change rather than only the marginal response distribution at an individual wave. It is computed using the common longitudinal evaluation set.

For a common question $q$, respondent $i$, and two adjacent waves $t$ and $t+1$, the human response transition is
\begin{equation}
y_{i,t}^{q}
\rightarrow
y_{i,t+1}^{q},
\end{equation}
while the corresponding model transition is
\begin{equation}
\hat{y}_{i,t}^{q}
\rightarrow
\hat{y}_{i,t+1}^{q}.
\end{equation}

We pool valid transitions across respondents and adjacent-wave pairs to construct empirical distributions over ordered response pairs:
\begin{equation}
P_h^{q}(a,b)
=
\Pr
\left(
y_t^{q}=a,
y_{t+1}^{q}=b
\right),
\end{equation}
\begin{equation}
P_m^{q}(a,b)
=
\Pr
\left(
\hat{y}_t^{q}=a,
\hat{y}_{t+1}^{q}=b
\right),
\end{equation}
where $(a,b)\in\mathcal{Y}_{q}\times\mathcal{Y}_{q}$ represents an ordered transition such as $1\!\rightarrow\!1$, $1\!\rightarrow\!2$, or $2\!\rightarrow\!1$. A transition is included only when both adjacent human responses are valid and both corresponding model responses can be parsed into valid options.

For each common question, we define the midpoint distribution
\begin{equation}
M^{q}
=
\frac{1}{2}
\left(
P_h^{q}+P_m^{q}
\right),
\end{equation}
and compute
\begin{equation}
\operatorname{JS}_{\mathrm{trans}}^{q}
=
\frac{1}{2}
\operatorname{KL}
\left(
P_h^{q}\,\|\,M^{q}
\right)
+
\frac{1}{2}
\operatorname{KL}
\left(
P_m^{q}\,\|\,M^{q}
\right).
\end{equation}

The reported transition-distribution JS divergence is the mean across eligible common questions:
\begin{equation}
\operatorname{TransitionJS}
=
\frac{1}{|\mathcal{Q}_{\mathrm{common}}|}
\sum_{q\in\mathcal{Q}_{\mathrm{common}}}
\operatorname{JS}_{\mathrm{trans}}^{q}.
\end{equation}

Lower values indicate that the model more closely reproduces the human population-level distribution of response persistence and change across adjacent waves.

\section{Additional Results}
\label{app:additional-results}

\subsection{Robustness Across Respondent Samples}
\label{app:sample-robustness}

To evaluate robustness to respondent sampling, we repeat the main experiments using three independently sampled respondent sets generated with seeds 42, 43, and 44. Each run contains 100 respondents, while all other experimental settings remain unchanged.
Table~\ref{tab:sample-robustness-main} reports the main-method comparison, while Table~\ref{tab:sample-robustness-ablation} reports the LifeMem ablation results. Both tables present the mean and standard deviation across the three respondent samples.

Table~\ref{tab:sample-robustness-main} shows that LifeMem remains the strongest method across respondent samples. The generally small standard deviations indicate that the main conclusions are not driven by a particular respondent sample.

Table~\ref{tab:sample-robustness-ablation} further shows that the full LifeMem model consistently outperforms both ablations. This result supports the complementary contributions of structured event memory and persistent parametric memory across different respondent samples.

\subsection{Valid Response Rates}
\label{app:valid-response-rates}

Table~\ref{tab:valid-response-rates} shows that all LLM-based methods achieve high valid response rates across models, generally exceeding those in the human survey data. The lower human rates reflect refusals, nonresponse, and other missing-value cases that naturally occur in real-world surveys. The consistently high rates across methods also indicate that the experimental framework reliably produces responses that can be mapped to the predefined answer options.

\section{Mechanism and Representation Analyses}
\label{app:mechanism-analyses}

\subsection{LoRA Visualization Procedure}
\label{app:lora-visualization-procedure}

To visualize the evolution of agent-specific parametric memory, we extract the LoRA parameters of every agent after each survey wave. For agent $i$ at wave $t$, all LoRA parameter tensors are flattened and concatenated into a single vector:
\begin{equation}
\boldsymbol{\theta}_{i,t}
=
\operatorname{concat}_{m\in\mathcal{M}}
\operatorname{vec}\!\left(
\boldsymbol{\Theta}_{i,t}^{(m)}
\right),
\end{equation}
where $\mathcal{M}$ denotes the consistently ordered set of LoRA tensors, including the corresponding \texttt{lora\_A} and \texttt{lora\_B} parameters across adapted modules. Using a fixed ordering ensures that each dimension has the same parameter identity across agents and waves.

We stack the vectors from all $N$ agents and $T$ waves into a single matrix:
\begin{equation}
\mathbf{X}
=
\begin{bmatrix}
\boldsymbol{\theta}_{1,1}^{\top} \\
\boldsymbol{\theta}_{1,2}^{\top} \\
\vdots \\
\boldsymbol{\theta}_{N,T}^{\top}
\end{bmatrix}.
\end{equation}
Each parameter dimension is then standardized globally across all agent--wave states:
\begin{equation}
X'_{j,d}
=
\frac{X_{j,d}-\mu_d}
{\sigma_d+\epsilon},
\end{equation}
where $\mu_d$ and $\sigma_d$ are the mean and standard deviation of dimension $d$ over all rows of $\mathbf{X}$, and $\epsilon$ is a small constant for numerical stability. This shared normalization prevents dimensions with larger numerical scales from dominating the projection.

We fit a single two-dimensional PCA projection to the globally standardized matrix:
\begin{equation}
\mathbf{Z}
=
\operatorname{PCA}_{2}(\mathbf{X}').
\end{equation}
The resulting point $\mathbf{z}_{i,t}\in\mathbb{R}^{2}$ represents the LoRA state of agent $i$ after wave $t$. For each agent, states are connected in chronological order,
\begin{equation}
\mathbf{z}_{i,1}
\rightarrow
\mathbf{z}_{i,2}
\rightarrow
\cdots
\rightarrow
\mathbf{z}_{i,T},
\end{equation}
forming a longitudinal parameter trajectory. Point colors indicate survey waves, while connecting lines identify states belonging to the same agent. 

\subsection{Additional LoRA Visualizations}
\label{app:additional-lora-visualizations}

\begin{figure}[htbp]
    \centering
    \includegraphics[width=\columnwidth]{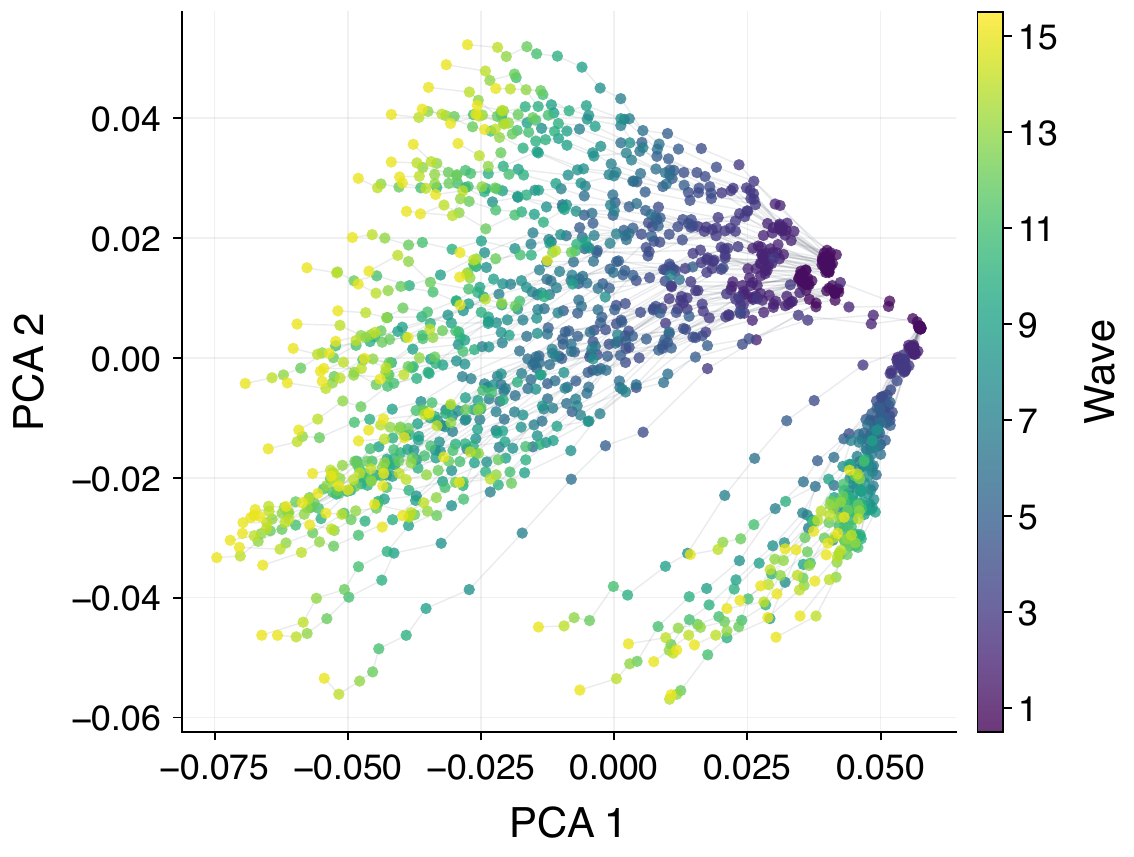}
    \caption{PCA visualization of agent-specific LoRA states learned by LifeMem for 100 agents on Understanding Society with Ministral-3-8B-Instruct-2512. Points represent agent states after each survey wave, and connected trajectories trace the same agent over time. Colors indicate survey waves.}
    \label{fig:lora-us-mistral}
\end{figure}

\begin{figure}[htbp]
    \centering
    \includegraphics[width=\columnwidth]{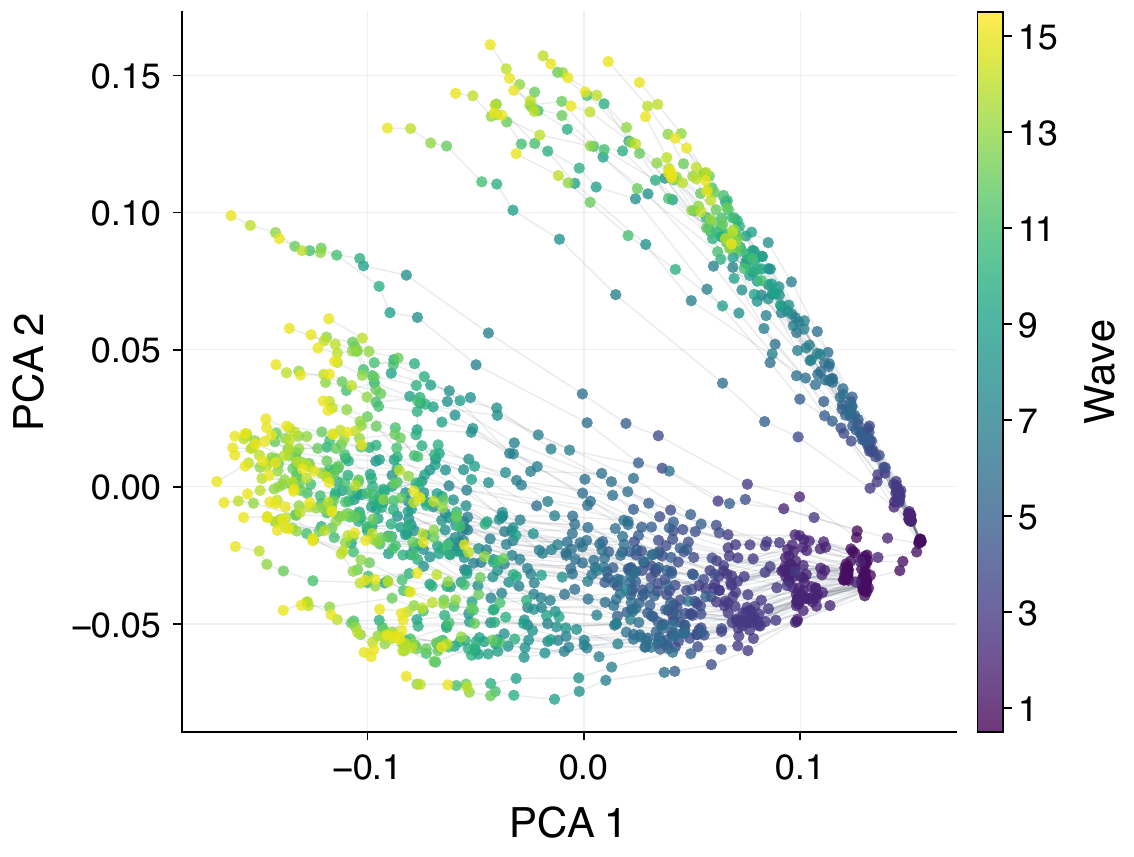}
    \caption{PCA visualization of agent-specific LoRA states learned by LifeMem for 100 agents on Understanding Society with Qwen3.5-9B. Points represent agent states after each survey wave, and connected trajectories trace the same agent over time. Colors indicate survey waves.}
    \label{fig:lora-us-qwen}
\end{figure}

Figures~\ref{fig:lora-us-mistral} and~\ref{fig:lora-us-qwen} extend the \emph{Evolution of Agent-Specific LoRA States} analysis in the main paper, which presents the Understanding Society results with Llama-3.1-8B-Instruct. Here, we provide the corresponding visualizations for the remaining backbone models: Ministral-3-8B-Instruct-2512 and Qwen3.5-9B on Understanding Society. Each point represents an agent's LoRA state after a survey wave, while connected points trace the same agent over time. Together, these figures show whether the longitudinal evolution of agent-specific parametric states observed in the main-paper example also appears across other backbone models.

\section{Retriever Choice}
\label{app:retriever-choice}

We compare several retrieval backends for \textsc{Event RAG} on Understanding Society with Ministral-3-8B-Instruct-2512. The evaluated retrievers include BM25, \texttt{all-MiniLM-L6-v2}, \texttt{E5-large-v2}, \texttt{BGE-M3}, and \texttt{BGE-M3} followed by \texttt{BGE-Reranker-v2-M3}. We additionally report LifeMem, which uses \texttt{all-MiniLM-L6-v2} for structured event retrieval while maintaining an agent-specific parametric memory.

Table~\ref{tab:retriever-choice} shows that retriever choice affects \textsc{Event RAG}. \texttt{BGE-M3} with reranking performs best among its variants on Understanding Society.
LifeMem achieves lower values across all metrics while using the same \texttt{all-MiniLM-L6-v2} retriever, suggesting that its gains are not solely due to retriever choice.

\begin{table*}[t]
\centering
\setlength{\tabcolsep}{4pt}
\begin{tabular}{l|l|ccc}
\toprule
\textbf{Category}
&
\textbf{Method}
&
KL Divergence $\downarrow$&
Within-Group Gap $\downarrow$&
Entropy Gap $\downarrow$
\\
\midrule

\multicolumn{5}{c}{\textbf{\textsc{Llama-3.1-8B-Instruct}}}\\
\midrule

\multirow{2}{*}{\textit{Static Conditioning}}
& Direct
& $15.317_{\pm 0.144}$
& $0.552_{\pm 0.004}$
& $0.698_{\pm 0.008}$
\\
& Profile
& $7.234_{\pm 0.072}$
& $0.376_{\pm 0.006}$
& $0.486_{\pm 0.011}$
\\
\multirow{2}{*}{\textit{Diversity-Oriented Prompting}}
& Multilingual
& $11.653_{\pm 0.074}$
& $0.341_{\pm 0.004}$
& $0.472_{\pm 0.004}$
\\
& Anti-Stereotype
& $8.372_{\pm 0.197}$
& $0.401_{\pm 0.006}$
& $0.515_{\pm 0.004}$
\\
\multirow{3}{*}{\textit{Non-Parametric Memory}}
& SimVBG
& $5.950_{\pm 0.315}$
& $0.345_{\pm 0.008}$
& $0.454_{\pm 0.017}$
\\
& Full History
& $4.658_{\pm 0.342}$
& $0.311_{\pm 0.008}$
& $0.388_{\pm 0.011}$
\\
& Event RAG
& $5.068_{\pm 0.209}$
& $0.308_{\pm 0.011}$
& $0.393_{\pm 0.008}$
\\
\textit{Control Baseline}
& Random Event
& $5.443_{\pm 0.125}$
& $0.341_{\pm 0.003}$
& $0.446_{\pm 0.007}$
\\
\rowcolor{gray!20}
\textit{Proposed Method}
& LifeMem
& $\mathbf{3.320}_{\pm 0.202}$
& $\mathbf{0.280}_{\pm 0.008}$
& $\mathbf{0.371}_{\pm 0.009}$
\\
\midrule
\multicolumn{5}{c}{\textbf{\textsc{Ministral-3-8B-Instruct-2512}}}\\
\midrule

\multirow{2}{*}{\textit{Static Conditioning}}
& Direct
& $15.408_{\pm 0.083}$
& $0.561_{\pm 0.004}$
& $0.713_{\pm 0.006}$
\\
& Profile
& $6.065_{\pm 0.674}$
& $0.346_{\pm 0.015}$
& $0.437_{\pm 0.022}$
\\
\multirow{2}{*}{\textit{Diversity-Oriented Prompting}}
& Multilingual
& $13.114_{\pm 0.070}$
& $0.393_{\pm 0.002}$
& $0.532_{\pm 0.003}$
\\
& Anti-Stereotype
& $8.243_{\pm 0.501}$
& $0.381_{\pm 0.012}$
& $0.490_{\pm 0.014}$
\\
\multirow{3}{*}{\textit{Non-Parametric Memory}}
& SimVBG
& $4.744_{\pm 0.272}$
& $0.303_{\pm 0.021}$
& $0.394_{\pm 0.020}$
\\
& Full History
& $3.462_{\pm 0.623}$
& $0.259_{\pm 0.041}$
& $0.321_{\pm 0.033}$
\\
& Event RAG
& $4.128_{\pm 0.237}$
& $0.273_{\pm 0.014}$
& $0.334_{\pm 0.008}$
\\
\textit{Control Baseline}
& Random Event
& $4.467_{\pm 0.226}$
& $0.307_{\pm 0.008}$
& $0.393_{\pm 0.010}$
\\
\rowcolor{gray!20}
\textit{Proposed Method}
& LifeMem
& $\mathbf{1.850}_{\pm 0.144}$
& $\mathbf{0.207}_{\pm 0.020}$
& $\mathbf{0.251}_{\pm 0.014}$
\\
\midrule
\multicolumn{5}{c}{\textbf{\textsc{Qwen3.5-9B}}}\\
\midrule

\multirow{2}{*}{\textit{Static Conditioning}}
& Direct
& $16.041_{\pm 0.044}$
& $0.551_{\pm 0.004}$
& $0.703_{\pm 0.005}$
\\
& Profile
& $5.431_{\pm 0.159}$
& $0.313_{\pm 0.029}$
& $0.394_{\pm 0.034}$
\\
\multirow{2}{*}{\textit{Diversity-Oriented Prompting}}
& Multilingual
& $11.029_{\pm 0.034}$
& $0.341_{\pm 0.003}$
& $0.469_{\pm 0.003}$
\\
& Anti-Stereotype
& $6.323_{\pm 0.348}$
& $0.312_{\pm 0.026}$
& $0.400_{\pm 0.024}$
\\
\multirow{3}{*}{\textit{Non-Parametric Memory}}
& SimVBG
& $4.839_{\pm 0.306}$
& $0.304_{\pm 0.013}$
& $0.384_{\pm 0.011}$
\\
& Full History
& $3.488_{\pm 0.251}$
& $0.273_{\pm 0.024}$
& $0.329_{\pm 0.014}$
\\
& Event RAG
& $3.659_{\pm 0.200}$
& $0.248_{\pm 0.020}$
& $0.296_{\pm 0.010}$
\\
\textit{Control Baseline}
& Random Event
& $3.938_{\pm 0.133}$
& $0.287_{\pm 0.014}$
& $0.356_{\pm 0.013}$
\\
\rowcolor{gray!20}
\textit{Proposed Method}
& LifeMem
& $\mathbf{2.622}_{\pm 0.140}$
& $\mathbf{0.228}_{\pm 0.021}$
& $\mathbf{0.276}_{\pm 0.016}$
\\
\bottomrule
\end{tabular}
\caption{Robustness across respondent samples on Understanding Society. Results report means across three independently sampled respondent sets generated with seeds 42, 43, and 44, with the corresponding $\pm$ standard deviations shown as subscripts. Each sample contains 100 respondents. Within-Group Gap and Entropy Gap denote within-group pairwise distance gap and normalized entropy gap. Lower values are better for all metrics. Bold marks the best mean result for each model and metric; LifeMem is shaded light gray.}
\label{tab:sample-robustness-main}
\end{table*}

\begin{table*}[t]
\centering
\setlength{\tabcolsep}{4pt}
\begin{tabular}{l|ccc}
\toprule
\textbf{Method}
&
KL Divergence $\downarrow$&
Within-Group Gap $\downarrow$&
Entropy Gap $\downarrow$
\\
\midrule

\multicolumn{4}{c}{\textbf{\textsc{Llama-3.1-8B-Instruct}}}\\
\midrule

\rowcolor{gray!20}
LifeMem (Full)
& $\mathbf{3.320}_{\pm 0.202}$
& $\mathbf{0.280}_{\pm 0.008}$
& $\mathbf{0.371}_{\pm 0.009}$
\\
LifeMem without Parametric Memory
& $4.744_{\pm 0.318}$
& $0.308_{\pm 0.016}$
& $0.402_{\pm 0.014}$
\\
LifeMem without Structured Memory
& $4.313_{\pm 0.413}$
& $0.317_{\pm 0.018}$
& $0.418_{\pm 0.013}$
\\
\midrule
\multicolumn{4}{c}{\textbf{\textsc{Ministral-3-8B-Instruct-2512}}}\\
\midrule

\rowcolor{gray!20}
LifeMem (Full)
& $\mathbf{1.850}_{\pm 0.144}$
& $\mathbf{0.207}_{\pm 0.020}$
& $\mathbf{0.251}_{\pm 0.014}$
\\
LifeMem without Parametric Memory
& $3.680_{\pm 0.206}$
& $0.267_{\pm 0.012}$
& $0.329_{\pm 0.010}$
\\
LifeMem without Structured Memory
& $2.268_{\pm 0.152}$
& $0.236_{\pm 0.011}$
& $0.300_{\pm 0.012}$
\\
\midrule
\multicolumn{4}{c}{\textbf{\textsc{Qwen3.5-9B}}}\\
\midrule

\rowcolor{gray!20}
LifeMem (Full)
& $\mathbf{2.622}_{\pm 0.140}$
& $\mathbf{0.228}_{\pm 0.021}$
& $\mathbf{0.276}_{\pm 0.016}$
\\
LifeMem without Parametric Memory
& $3.619_{\pm 0.247}$
& $0.253_{\pm 0.019}$
& $0.304_{\pm 0.014}$
\\
LifeMem without Structured Memory
& $3.521_{\pm 0.270}$
& $0.274_{\pm 0.013}$
& $0.342_{\pm 0.015}$
\\
\bottomrule
\end{tabular}
\caption{Robustness of the LifeMem ablations across respondent samples. Results report means across three independently sampled respondent sets generated with seeds 42, 43, and 44, with the corresponding $\pm$ standard deviations shown as subscripts. Within-Group Gap and Entropy Gap denote within-group pairwise distance gap and normalized entropy gap. All metrics are lower-is-better. Bold marks the best mean result for each model and metric; LifeMem (Full) is shaded light gray.}
\label{tab:sample-robustness-ablation}
\end{table*}

\begin{table*}[t]
\centering
\setlength{\tabcolsep}{4pt}
\resizebox{\textwidth}{!}{
\begin{tabular}{lccc|lccc}
\toprule
\multicolumn{8}{c}{\textbf{Understanding Society}} \\
\midrule
\textbf{Method} & \textbf{Llama-8B} & \textbf{Ministral-8B} & \textbf{Qwen3.5-9B} &
\textbf{Method} & \textbf{Llama-8B} & \textbf{Ministral-8B} & \textbf{Qwen3.5-9B} \\
\midrule
\textbf{Human} & \multicolumn{3}{c|}{\textbf{64.31\%}} & & & & \\
\midrule
Direct & 91.33\% & 90.87\% & 92.27\% & Profile & 91.58\% & 91.14\% & 93.04\% \\
Multilingual & 89.07\% & 91.01\% & 92.00\% & Anti-Stereotype & 91.54\% & 90.97\% & 93.29\% \\
SimVBG & 91.60\% & 91.71\% & 93.02\% & Full History & 91.73\% & 91.35\% & 93.34\% \\
Event RAG & 91.63\% & 91.04\% & 93.15\% & Random Event & 91.72\% & 91.05\% & 93.11\% \\
LifeMem & 92.28\% & 91.59\% & 93.34\% &  &  &  &  \\
\bottomrule
\end{tabular}
}
\caption{Valid response rates across methods and backbone models on Understanding Society. Two methods are shown per row, read from left to right; the final row contains LifeMem alone. Llama-8B, Ministral-8B, and Qwen3.5-9B denote Llama-3.1-8B-Instruct, Ministral-3-8B-Instruct-2512, and Qwen3.5-9B, respectively. A response is valid when it can be mapped to a predefined answer option. Higher values indicate fewer missing or unparsable responses.}
\label{tab:valid-response-rates}
\end{table*}

\begin{table*}[t]
\centering
\begin{tabular}{l|l|lll}
\toprule
\textbf{Method}
&
\textbf{Retriever}
&
KL Divergence $\downarrow$&
Within-Group Gap $\downarrow$&
Entropy Gap $\downarrow$
\\
\midrule

\multirow{5}{*}{Event RAG}
& BM25
& 4.237 & 0.290 & 0.356
\\
& all-MiniLM-L6-v2
& 3.845 & 0.281 & 0.339
\\
& E5-large-v2
& 4.466 & 0.299 & 0.355
\\
& BGE-M3
& 4.334 & 0.289 & 0.341
\\
& BGE-M3 + BGE-Reranker-v2-M3
& 3.597 & 0.277 & 0.322
\\
\midrule

\rowcolor{gray!20}
LifeMem
& all-MiniLM-L6-v2
& \textbf{1.966} & \textbf{0.228} & \textbf{0.266}
\\
\bottomrule
\end{tabular}
\caption{Effect of retriever choice on \textsc{Event RAG} and LifeMem with Ministral-3-8B-Instruct-2512. Lower values are better for all metrics. Bold marks the best overall result, and LifeMem is shaded light gray.}
\label{tab:retriever-choice}
\end{table*}

\end{document}